# Collaborative Management for Chronic Diseases and Depression: A Double Heterogeneity-based Multi-Task Learning Method

Yidong Chai, Ph.D.

City University of Hong Kong

Email: yidongchai@gmail.com

Haoxin Liu, Ph.D.

University of Science and Technology of China

Email: hxliu1215@ustc.edu.cn

Jiaheng Xie, Ph.D.

University of Delaware

Email: jxie@udel.edu

Chaopeng Wang

Hefei University of Technology

Email: wangcp@mail.hfut.edu.cn

Xiao Fang, Ph.D.

University of Delaware

Email: xfang@udel.edu

# Collaborative Management for Chronic Diseases and Depression: A Double Heterogeneity-based Multi-Task Learning Method

**Abstract**

Wearable sensor technologies and deep learning are transforming healthcare management. Yet, most health sensing studies focus narrowly on physical chronic diseases. This overlooks the critical need for joint assessment of comorbid physical chronic diseases and depression, which is essential for collaborative chronic care. We conceptualize multi-disease assessment—including both physical diseases and depression—as a multi-task learning (MTL) problem, where each disease assessment is modeled as a task. This joint formulation leverages inter-disease relationships to improve accuracy, but it also introduces the challenge of double heterogeneity: chronic diseases differ in their manifestation (disease heterogeneity), and patients with the same disease show varied patterns (patient heterogeneity). To address these issues, we first straightforwardly adopt existing techniques and propose a base method. Given the limitations of the base method, we further propose an Advanced Double Heterogeneity-based Multi-Task Learning (ADH-MTL) method that improves the base method through three innovations: (1) group-level modeling to support new patient predictions, (2) a novel decomposition strategy to reduce model complexity, and (3) a novel Bayesian network that explicitly captures dependencies while balancing similarities and differences across model components. Empirical evaluations on real-world wearable sensor data demonstrate that ADH-MTL significantly outperforms existing baselines, and each of its innovations proves to be effective. This study contributes to health information systems by offering a principled computational solution for integrated physical–mental healthcare and provides design principles for advancing collaborative chronic disease management across the pre-treatment, treatment, and post-treatment phases.

## 1. Introduction

Chronic diseases pose critical societal challenges, causing severe threats to individual well-being, public health systems, and overall socioeconomic stability. From 2011 to 2030, the cumulative global healthcare expenditure on chronic diseases is projected to reach $47 trillion (Hacker 2024). This burden is further intensified by the progressive nature of these diseases, as patients often develop comorbidities – 30% of adults globally now live with two or more chronic diseases (Chowdhury et al. 2023). Effective management of comorbid chronic diseases is crucial for improving patient outcomes, mitigating disease progression, and generating significant social benefits, thus gaining prominence in the literature (Brohman et al. 2020, Aron and Pathak 2021).

Effectively managing comorbid chronic diseases requires not only assessing each disease individually but also modeling their interactions. Multi-disease assessment has emerged as a promising method to comprehensively understand patient health by capturing inter-disease relationships, ultimately improving healthcare outcomes (S. Xie et al. 2021). Given these advantages, a growing number of IS studies have begun to explore novel methods to jointly assess multiple diseases (Lin et al. 2017, Wang et al. 2024). A notable example is Lin et al. (2017). They developed a Bayesian logistic regression model utilizing electronic health records (EHRs) to assess multiple chronic diseases, including diabetes, stroke, acute myocardial infarction, and acute renal failure. However, existing studies alike (Lin et al. 2017, Wang et al. 2024) have two main limitations. First, they mainly focus on physical chronic diseases. While this focus has improved understanding of physical comorbidities, it neglects the critical role of depression in managing chronic diseases. This is problematic because the sustained symptom burden and complex self-management often trigger depression, and thus depression is highly prevalent among patients with physical chronic diseases. Moreover, depression lowers motivation, impairs cognition, and reduces treatment adherence, all undermining effective disease management (Mezuk et al. 2008, Lippi et al. 2009). Hence, collaborative care requires the integrated management of depression within chronic disease care. Second, the long-term progression and recovery of chronic diseases require continuous real-time assessment. However, the existing methods primarily rely on EHRs or medical imaging (Lin et al. 2017, Wang et al. 2024), which are episodic and clinic-based, making continuous real-time assessment costly and difficult.

Advancements in wearable sensor technologies are reshaping chronic disease management. As Bardhan et al. (2020) noted, "*wearable sensors and home devices can play a pivotal role not only in monitoring the status of disease patients and predicting adverse events before they occur but also in preventing the onset of diseases.*" For physical chronic diseases, many IS studies (Yu et al. 2022, 2024) have shown that wearable sensors can effectively monitor behavioral symptoms such as distinct gait patterns to assess disease status. Although depression is a mental disorder, medical research has shown that it often manifests through behavioral symptoms such as slowed movement and fatigue (NHS 2023), making it detectable through wearable sensors. Hence, there is an urgent need to jointly assess comorbid chronic diseases and depression to support effective collaborative care. *Our first objective is to develop a wearable sensor-based method for the joint assessment of comorbid physical chronic diseases and depression.*

The most suitable approach for this objective is multi-task learning (MTL), which models each disease as a separate task and leverages inter-disease relationships to improve prediction. While MTL addresses disease heterogeneity, most existing methods assume population-level models and ignore patient heterogeneity—the distinct patterns patients exhibit even within the same disease. For example, diabetic patients may develop distinct gait abnormalities: some exhibit cautious, short strides with high step frequency due to nerve damage from chronic hyperglycemia, while others show intermittent walking patterns from fatigue requiring frequent rests (Fritschi and Quinn 2010, Hulshof et al. 2024). Patient heterogeneity limits the effectiveness of current MTL methods in delivering accurate and personalized disease assessments. *Our second objective is to design an MTL-based assessment method that jointly tackles the double heterogeneity at both the disease and patient levels.*

This study begins by applying existing techniques to this problem and proposes a Base Double Heterogeneity-based Multi-Task Learning (BDH-MTL) method. While not being our main contribution, BDH-MTL addresses the double heterogeneity by creating a personalized assessment model for each patient-disease combination with two phases: structure design and model learning. In the structure design phase, given the unstructured nature of wearable sensor data, we design a deep learning-based assessment model composed of a feature extraction component and a prediction component. In the second phase, we use disease-patient relationships, a four-dimensional matrix (e.g., (Disease A-Patient B): (Disease C-Patient D)), to learn the parameters of each assessment model.

BDH-MTL establishes the foundation for effective personalized learning to address the problem of double heterogeneity. However, this base model is limited by four challenges: 1) its individual-level modeling is not applicable to new patients, 2) its four-dimensional relationship matrix is overly complex and difficult to learn, 3) its lack of explicit dependency modeling among relationships, parameters, and performance impairs learning, and 4) its lack of accounting for both the differences and similarities between model components hinders performance. These challenges motivate us to propose a novel, *advanced* method, named Advanced Double Heterogeneity-based Multi-Task Learning (ADH-MTL). ADH-MTL includes three key designs: 1) group-level modeling, 2) a novel four-dimensional relationship decomposition, and 3) a novel Bayesian network. Group-level modeling clusters patients and builds group-specific personalized models. For a new patient, we assign them to the most suitable group and apply its corresponding model, thus addressing the first challenge. We then decompose the complex four-dimensional relationship matrix into two two-dimensional matrices. This decomposition renders the learning process much simpler, thereby addressing the second challenge. Next, we introduce a novel Bayesian network that explicitly models dependencies among relationships, model parameters, and performance, allowing us to learn assessment models with high performance, thereby addressing the third challenge. Moreover, in our Bayesian network, we define two separate but dependent relationships for the two components of the assessment models, thereby addressing the fourth challenge.

The contributions of this study are two-fold. First, it addresses a significant gap in the IS literature on health sensing. Existing research overlooks the chronic disease comorbidity and the interaction between physical diseases and depression. Positioned in the computational design science paradigm, this study develops novel IT artifacts to address this gap using wearable sensor data. Second, from a methodological perspective, this work addresses the double heterogeneity problem (i.e., disease and patient heterogeneity) in MTL. We first apply existing techniques to develop BDH-MTL to explicitly tackle this problem. To address the limitations of BDH-MTL, we further propose a novel ADH-MTL, which introduces a novel decomposition of a complex four-dimensional relationship matrix to reduce complexity and improve learning effectiveness. ADH-MTL also includes a novel Bayesian network to explicitly model the intricate dependencies arising from simultaneously considering the double heterogeneity. Hence, this study contributes to the MTL field by introducing novel methods to address the problem of double heterogeneity.

## 2. Literature Review

### 2.1 Health Information Technology Studies in IS

Advances in health information technology (HIT), including health information systems (HIS), patient portals, online health communities (OHC), and electronic health records (EHRs), have expanded the role of IS and data analytics in healthcare management (Baird et al. 2020). Meanwhile, computational methods such as AI for healthcare management have grown increasingly popular within the IS discipline.

According to Bardhan et al. (2020), computational HIT research has generally progressed along three streams. The first stream primarily leverages structured clinical data such as EHRs to address health issues, including adverse event prediction (Lin and Fang 2021) and hospital readmission forecasting (Xie et al. 2021a). The second stream focuses on data from online health communities to examine topics, such as patient education (Liu et al. 2019), drug safety surveillance (Xie et al. 2021b), and medication adherence (Xie et al. 2022). The third stream builds on advances in mobile apps and wearable sensors. As Bardhan et al. (2020) highlight, wearable sensors offer promising capabilities for monitoring chronic diseases, predicting critical events, and enabling early interventions. This highlights the urgent need to push forward the third stream of HIT research by creating novel wearable sensor-based healthcare IT artifacts. Accordingly, leading IS journals have recently published a growing body of work in this stream, as summarized in Table A.1 of Appendix A. For example, Zhu et al. (2020) propose a novel dual-kernel CNN combined with deep transfer learning to identify activities of daily living from wearable sensor data in senior care settings. Yu et al. (2022) propose an attention-based neural network that uses mobile sensor signals to assess Parkinson's disease severity.

Positioned within the third stream, our study aims to develop novel predictive methods using wearable sensors for effective chronic disease management. Meanwhile, from Table A.1, we also identify two important gaps that remain to be unaddressed. First, many studies focus on a single disease, overlooking the comorbidity of chronic diseases that calls for joint assessment. Although the broader literature has explored multi-disease settings (for example, using Bayesian logistic regression with EHRs (Lin et al. 2017) and using Hidden Markov Models combined with Logistic Regression with EHRs (Ben-Assuli and Padman 2020)), the complexity and unstructured nature of wearable sensor data make the existing methods unsuitable for sensor-based contexts. Second, prior research has not addressed the collaborative management of physical chronic diseases and depression. Depression is highly prevalent among patients with chronic diseases and can substantially hinder recovery (J. Katon 2011). Studies show that integrating depression into chronic disease management improves clinical outcomes (Katon et al. 2010), whereas neglecting it worsens health outcomes, increases hospitalization and mortality, and raises healthcare costs

(J. Katon 2011). Thus, there is a critical need for collaborative care that integrates the management of comorbid chronic diseases and depression to promote holistic well-being. Multi-disease assessment is essential for such care, as it enables targeted interventions tailored to the severity of each disease. Our study focuses on leveraging wearable sensor technologies to jointly assess comorbid chronic diseases and depression.

**2.2 Multi-Task Learning**

Technically, multi-disease assessment can be viewed as MTL, wherein the assessment of each disease constitutes an individual task. Unlike single-task learning, which learns a model for a single specific task, MTL enables knowledge sharing across related tasks, yielding better performance than learning each task independently (Ruder 2017). MTL studies are divided into two types based on data structure: those where each task has its own separate dataset, and those where all tasks share a single dataset with each instance being labeled for every task. In our multi-disease assessment context, each disease is assessed based on the same input instance (i.e., patient's data). Therefore, it falls into the latter type of MTL, which can be further divided into two approaches: feature sharing and parameter sharing.

Feature sharing involves learning a common set of features across all tasks. The original, naive approach shares the whole set of features (Kendall et al. 2018). However, this strict sharing design can limit model performance, as it fails to capture the distinct characteristics of different tasks, and thus cannot address the problem of task heterogeneity (referred to as disease heterogeneity in our context). Subsequent research improved upon this in four directions: 1) weighted aggregation of shared features (Zhang et al. 2021), 2) feature selection from shared features (Yin et al. 2024), 3) simultaneously learning task-specific features (Tan et al. 2023), and 4) dynamic module selection (Rahimian et al. 2023). Weighted aggregation methods first learn a common set of features and then apply task-specific aggregation, like the attention mechanism, to create a personalized set of features for each task. Feature selection methods select a personalized subset from the common set for each task. A key limitation of these two approaches is their fundamental dependence on the initial common set of features. This dependence inherently limits personalization, as the shared features are often inadequate for capturing each task's unique nuances. The third approach learns both shared features and task-specific features. Analyses for each task are then made by combining its task-specific features with the shared features. A key limitation is that they not only need to learn useful features but also to determine the optimal assignment for each feature, whether it should be shared or task-specific. The distinction between these two types of features is often ambiguous. A feature may be beneficial for several tasks, but to varying degrees. This ambiguity complicates the feature learning process, often resulting in instability and reduced performance. The fourth approach bypasses the practice of a monolithic set of shared features. Instead, it obtains personalized features by dynamically selecting pathways composed of fine-grained modules (e.g., layers) within the model structure. Since each module extracts distinct features, different pathways yield different features. Meanwhile, since tasks may share some modules in their pathways, they also partially share their features. This approach can effectively overcome the limitations of a rigid shared feature set. However, it tends to cause a Matthew effect, where frequently chosen components get more training and become increasingly dominant. This feedback loop often leads to model collapse, with only a few modules being learned.

Parameter sharing addresses task heterogeneity by building a separate, personalized model for each task. Based on how information is shared across tasks, the existing methods can be broadly categorized into three types: 1) distribution-based (Li et al. 2014), 2) matrix decomposition-based (Lin et al. 2017), and 3) relationship-based approaches (Nishi et al. 2024). The distribution-based approach treats model parameters as random variables from distributions and then models their joint distribution across tasks. By learning the joint distribution of parameters, this approach coordinates across tasks to facilitate information sharing. For example, Li et al. (2014) represent the parameters of each SVM model as columns of a matrix and assume that these parameters follow a matrix-variate normal distribution. They then optimize this distribution, from which the parameters of each model (corresponding to a column of the matrix) are derived. The second, matrix decomposition-based approach decomposes the parameter matrix (each column corresponding to the parameter of a task-specific model) into a shared part and a task-specific part, which are then learned simultaneously through optimization. The limitation of these two approaches is that they are mainly designed for classical machine learning models (e.g., SVM or logistic regression), whose parameters are vector-based and can thus be stacked into a matrix across multiple models. However, for deep learning models, which are more effective in our context to analyze wearable sensor data, the large number of parameters makes them unsuitable. The third approach, the relationship-based approach, leverages task relationships to learn task-specific model parameters and has been widely applied in both classical machine learning and deep learning. This approach can be further divided into two types: similarity-regularization-based (Sang et al. 2024) and parameter-aggregation-based approaches (Hervella et al. 2024). The similarity-regularization-based approach promotes information sharing by imposing similarity constraints, such as L2 norms, on task-specific parameters for models of similar tasks. The limitation is that regularization is typically imposed on every pair of selected layers across all model pairs. With $I$ tasks (and thus $I$ models), each containing $L$ layers subject to regularization, the total number of regularizations scales as $L \cdot \binom{I}{2}$. Moreover, the importance of each regularization is often subjective and hard to determine, and this challenge becomes more severe as the number of regularizations grows. The parameter-aggregation-based approach, in contrast, aggregates each task's model by taking a weighted average of the current parameters of all models, with the weights determined by the inter-task relationships. An iterative process is performed, where the aggregated parameters from the previous round are used to aggregate new parameters in the next round. This approach has frequently yielded the best results and has grown increasingly popular in recent years. Accordingly, we focus on it, with the major studies summarized in Table B.1 of Appendix B. For instance, Sen and Borcea (2024) calculate inter-task similarity using the cosine similarity of model parameters. Tasks with low similarity are removed, and the remaining scores are normalized to produce weights for aggregating model parameters. The aggregated parameters are then used to update models in the next round. Their method performs well on tasks including face attribute classification and semantic segmentation. Similarly, Tsouvalas et al. (2025) calculate task vectors as the difference between each task's updated and pre-updated parameters. These vectors are used to measure inter-task similarity and assign weights for parameter aggregation. Their methods perform well on tasks such as scene and land-use classification, fine-grained object classification, digit and traffic-sign recognition, and texture classification.

However, a key limitation in the current MTL studies, including the parameter-aggregation-based approach, is that existing methods mainly focus on task heterogeneity (or disease heterogeneity in our context), while failing to account for *entity heterogeneity*, which refers to patient heterogeneity in our context. In other words, they operate at the population level, with the same models applied to all entities (i.e., patients). For instance, Cui and Mitra (2024) create a model for multiple tasks, including renal failure, acute cerebrovascular disease, acute cardiac dysrhythmias, and kidney disease, to address the task heterogeneity. However, the same set of models is applied to all entities with different ages, genders, ethnicities, marital statuses, and lengths of hospital stays. As a result, they overlook the distinct and sometimes conflicting patterns exhibited by patients with the same disease. For example, patients with depression may experience insomnia (difficulty falling asleep) or hypersomnia (prolonged daytime sleep and inactivity) (Nutt et al. 2008). Walking patterns vary among patients with depression: some become largely sedentary, while others maintain daily activity levels (Sloman et al. 1982, De Mello et al. 2013). Similarly, patients with cardiovascular disease often exhibit considerable heterogeneity: some develop heart failure, while others experience arrhythmia-related conditions such as palpitations or dizziness (Jafari et al. 2023, Wu et al. 2024). These heterogeneous manifestations underscore the limitations of population-level modeling commonly adopted in current MTL methods. Hence, novel methods are needed to jointly address both disease heterogeneity and patient heterogeneity to enable more accurate multi-disease assessment.

## 3. The Proposed Base Double Heterogeneity-based Multi-Task Learning (BDH-MTL) Method

### 3.1 Problem Formulation

For a patient $p$, let $\boldsymbol{y}_p = (y_{1,p}, \ldots, y_{d,p}, \ldots, y_{D,p})$ denote the patient's multi-disease status, where $y_{d,p} = 1$ indicates the presence of disease $d$, and $y_{d,p} = 0$ indicates its absence. $y_{1,p}$ to $y_{D-1,p}$ correspond to the physical chronic diseases, while $y_{D,p}$ corresponds to depression. The wearable sensor data is denoted as $\boldsymbol{X}_p^{\text{sensor}}$. We also include the patient's profile information, including demographic characteristics, family medical history, and obesity status, denoted as $\boldsymbol{X}_p^{\text{profile}}$. Collectively, the dataset is denoted as $\mathcal{D} = \left\{ \left( \boldsymbol{X}_p^{\text{sensor}}, \boldsymbol{X}_p^{\text{profile}}, \boldsymbol{y}_p \right) \middle| p = 1, 2, \ldots, P \right\}$, where $P$ denotes the number of patients. This study aims to learn

models from dataset $\mathcal{D}$ that jointly assesses the status of comorbid physical chronic diseases and depression for each patient, utilizing their wearable sensor data and profile information. Given the heterogeneity of both diseases and patients, we account for the double heterogeneity to enable more accurate assessments.

### 3.2 The Proposed Base Method

We first apply existing techniques and propose a base method, BDH-MTL, to address double heterogeneity. BDH-MTL builds a personalized assessment model for each patient-disease combination and leverages patient-disease relationships to learn model parameters. BDH-MTL serves as a foundation for understanding the challenges of double heterogeneity. These challenges, in turn, motivate the design of a more advanced method. We present the base method in this section and the advanced one in the next.

BDH-MTL consists of two major phases: 1) structure design, and 2) model learning. In the first phase, we design the structure of the assessment model to ensure that it can functionally operate, i.e., output probability scores indicating the likelihood that a patient has specific diseases, given wearable sensor data and profile information. The second phase learns the parameters of the assessment models.

#### 3.2.1 Designing the Structure of the Assessment Models

Given the unstructured nature of wearable sensor data, we employ deep learning models in our assessment model. Similar to existing deep learning models, our model consists of two main components: a feature extraction component and a prediction component. The feature extraction component is designed to generate useful vector-based features (also called representations) from the unstructured input. To capture both spatial and temporal information of wearable sensor data, we adopt the CNN-LSTM (Pallewar et al. 2024), where the CNN analyzes spatial information and the LSTM analyzes sequential information. Formally, we denote this process as $\boldsymbol{e}_p^{\text{sensor}} = \text{CNNLSTM}(\boldsymbol{X}_p^{\text{sensor}})$, where $\boldsymbol{e}_p^{\text{sensor}}$ represents the features extracted from the wearable sensor data $\boldsymbol{X}_p^{\text{sensor}}$. Meanwhile, we use an MLP to extract vector-based features from patients' profile information, denoted as $\boldsymbol{e}_p^{\text{profile}} = \text{MLP}_1(\boldsymbol{X}_p^{\text{profile}})$, where $\boldsymbol{e}_p^{\text{profile}}$ represents the corresponding extracted features. The prediction component, as in most deep learning models, is designed as an MLP. It maps the extracted features to a probability score that indicates the likelihood of the presence of a certain disease (e.g., the $d$-th disease). We denote this process as $\hat{y}_{d,p} = \text{MLP}_2\left(e_p^{\text{sensor}}; e_p^{\text{profile}}\right)$ where $\hat{y}_{d,p}$ is the predicted probability score. The overall structure is shown in Figure C.1 of Appendix C.

**3.2.2 Learning Personalized Assessment Models**

As is common practice, all models employ the same structure to facilitate parameter sharing, while their parameters differ. To tackle both disease heterogeneity and patient heterogeneity, BDH-MTL creates a personalized model for each patient-disease combination. Without loss of generality, we illustrate the assessment of disease $d$ for patient $p$. We create an assessment model denoted as $M_{d,p}$ with parameter $\boldsymbol{\theta}_{d,p}$. Next, we describe how to learn parameter $\boldsymbol{\theta}_{d,p}$.

The learning process is similar to existing parameter aggregation–based MTL studies and is guided by three key considerations. First, the learning of $\boldsymbol{\theta}_{d,p}$ should be guided by the performance of $M_{d,p}$. However, relying solely on this can lead to overfitting, necessitating the use of information from other models. This leads to the second consideration: the learning of parameter $\boldsymbol{\theta}_{d,p}$ should use the parameters of all related models and the relationships among them. For instance, if diseases $d$ and $d'$ are close and patients $p$ and $p'$ are close, then $\boldsymbol{\theta}_{d,p}$ should have a greater influence on learning $\boldsymbol{\theta}_{d',p'}$. Third, since the relationships among patient-disease combinations influence the learning and are difficult to determine manually, they should be learned from data. Accordingly, the learning involves the following steps.

<u>Step 1 (Initialization)</u>: We randomly initialize model parameters and model relationships. Since a model is created for each patient-disease combination, the relationships among the models are equal to the relationships among the patient-disease combinations. We denote $R_{(d,p),(d',p')}$ as the relationship between the combination of disease $d$ and patient $p$ and the combination of disease $d'$ and patient $p'$ (equivalently, between the corresponding models $M_{d,p}$ and $M_{d',p'}$). All the relationships are collectively denoted as $\boldsymbol{R}$. Note that $\boldsymbol{R}$ is a four-dimensional matrix of the form "(disease, patient) - (disease, patient)," where each element is indexed by four variables. For example, $R_{(d,p),(d',p')}$ has the indexes of $d, p, d', p'$.

Step 2 (Relationship-based Parameter Aggregation): We adopt the classic parameter aggregation technique, where the parameters of a focal model (e.g., $\boldsymbol{\theta}_{d,p}$) are updated via aggregation, i.e., a weighted average of all related models with the model relationships serving as the weights. Specifically, given current parameters $\{\widetilde{\boldsymbol{\theta}}_{d,p}|\forall d,p\}$, the parameter $\boldsymbol{\theta}_{d,p}$ is updated with Equation (1):

$$\boldsymbol{\theta}_{d,p} = \frac{\sum_{d'=1}^{D}\sum_{p'=1}^{P} R_{(d,p),(d',p')}\widetilde{\boldsymbol{\theta}}_{d',p'}}{\sum_{d'=1}^{D}\sum_{p'=1}^{P} R_{(d,p),(d',p')}}. \tag{1}$$

Step 3 (Performance-Guided Parameter Updating): We further update $\boldsymbol{\theta}_{d,p}$ to maximize model $M_{d,p}$'s performance. Let $L_{d,p}(\boldsymbol{\theta}_{d,p})$ denote the performance loss in assessing disease $d$ for patient $p$ with parameters $\boldsymbol{\theta}_{d,p}$. We update $\boldsymbol{\theta}_{d,p}$ to minimize $L_{d,p}(\boldsymbol{\theta}_{d,p})$ with a gradient descent algorithm (e.g., Adam):

$$\boldsymbol{\theta}_{d,p} \leftarrow \text{GradientDesecent}\big(L_{d,p}(\boldsymbol{\theta}_{d,p}), \boldsymbol{\theta}_{d,p}\big) \tag{2}$$

Note that Steps 2 and 3 are applied to update all models' parameters, i.e., $\{\boldsymbol{\theta}_{d,p}|\forall d,p\}$.

Step 4 (Performance-Guided Relationship Updating): We update the relationships $\boldsymbol{R}$ based on the performance across all diseases and patients with a gradient descent algorithm (e.g., Adam):

$$\boldsymbol{R} \leftarrow \text{GradientDesecent}\left(\sum_{d=1}^{D}\sum_{p=1}^{P} L_{d,p}(\boldsymbol{\theta}_{d,p}), \boldsymbol{R}\right) \tag{3}$$

After Step 4, a new training round begins from Step 2. The newly obtained parameters $\{\boldsymbol{\theta}_{d,p}|\forall d,p\}$ in Step 3 become $\{\widetilde{\boldsymbol{\theta}}_{d,p}|\forall d,p\}$ in the next round. This process is repeated iteratively until convergence. The pseudocode is in Appendix D. After learning parameters $\boldsymbol{\theta}_{d,p}$ for each model $M_{d,p}$, we obtain a personalized assessment model that addresses heterogeneity across both diseases and patients.

## 4. The Proposed Advanced Double Heterogeneity-based Multi-Task Learning (ADH-MTL) Method

BDH-MTL straightforwardly addresses the issue of double heterogeneity in multi-task learning. However, several challenges remain. First, individual-level modeling is not applicable to new patients. Since no existing assessment model is available for new patients, a new model must be created from scratch, which is costly and inconvenient. Second, the four-dimensional relationship matrix in BDH-MTL is complex and hard to learn. Third, BDH-MTL fails to explicitly capture dependencies among model relationships, model parameters, and performance. Since these three factors are interdependent, it is essential to explicitly model their relationships for effective parameter learning. Fourth, in BDH-MTL, the two components of the assessment model share the same relationship. However, they serve distinct purposes and may therefore exhibit different relationships. Yet, as parts of the same model for the same application (e.g., same disease), they share similarities. Hence, simultaneously accounting for both the differences and similarities when learning these relationships is the fourth challenge.

The four challenges above motivate us to propose an Advanced Double-Heterogeneity-based Multi-Task Learning (ADH-MTL) method. Like BDH-MTL, ADH-MTL consists of two phases: 1) model design and 2) model learning. In the first phase, it shares the same assessment model structure as BDH-MTL. The key improvements lie in the learning phase, which incorporates three key designs: 1) group-level modeling to address the first challenge, 2) four-dimensional relationship decomposition to address the second challenge, and 3) a novel Bayesian network to address the third and fourth challenges.

**4.1 First Design: Group-level Modeling**

In clinical practice, certain subpopulations exhibit similar patterns. For instance, elderly patients with depression often show loss of appetite (Hybels et al. 2012), while younger patients more commonly experience sleep disturbances, social withdrawal, and irregular digital activity (Crouse et al. 2021). Motivated by these observations, we propose clustering patients into $K$ groups based on their profile information. Patients in the same group are closely related, whereas those in different groups are more distant and may show heterogeneous patterns. Accordingly, we learn group-specific models. For a new patient, we identify their group from their profile and then apply the corresponding models for assessment.

Specifically, we apply the K-means to segment the patients into $K$ groups $\mathcal{G} = \{g_1, \ldots, g_k, \ldots, g_K\}$. We then train a group-specific model $M_{d,k}$ for disease $d$ and patients in group $k$. Each model $M_{d,k}$ is trained using the data corresponding to disease $d$ and patients in group $k$. For a new patient, we first determine their group $k'$, based on profile information, and then apply the set of models $\{M_{d,k'} | \forall d\}$ of group $k'$ to assess their disease status across multiple comorbid chronic diseases and depression.

## 4.2 Second Design: Decomposing Four-dimensional Matrix

Group-level modeling also reduces the number of models to $D \cdot K$, and simplifies the relationships into $R_{(d,k),(d',k')}$, denoting the relationship between "disease $d$, group $k$" and "disease $d'$, group $k'$." However, the relationships remain four-dimensional, making them hard to learn. $R_{(d,k),(d',k')}$ is closely related to both the inter-disease and the inter-group relationships. We model it as a linear combination of the inter-disease relationship $R^{\text{d}}_{(d,d')}$ and the inter-group relationship $R^{\text{g}}_{(k,k')}$: $R_{(d,k),(d',k')} \approx aR^{\text{d}}_{(d,d')} + (1-a)R^{\text{g}}_{(k,k')}$, where $a \in [0,1]$ is a learnable parameter. Then, we prove that the aggregation in Equation (1) is simplified to that in Equation (4) (derivation in Appendix E).

$$\boldsymbol{\theta}_{d,k} = a \frac{1}{\sum_{d'=1}^{D} R^{\text{d}}_{(d,d')}} \sum_{d'=1}^{D} \left( R^{\text{d}}_{(d,d')} \widetilde{\boldsymbol{\theta}}_{d',k} \right) + (1-a) \frac{1}{\sum_{k'=1}^{K} R^{\text{g}}_{(k,k')}} \sum_{k'=1}^{K} \left( R^{\text{g}}_{(k,k')} \widetilde{\boldsymbol{\theta}}_{d,k'} \right) \quad (4)$$

$\{\widetilde{\boldsymbol{\theta}}_{d,k} | \forall d, k\}$ denotes the current parameters. With Equation (4), $\boldsymbol{\theta}_{d,k}$ is aggregated based on a relative weight $a$ and two two-dimensional matrices, $\boldsymbol{R}^{\text{d}}$ and $\boldsymbol{R}^{\text{g}}$. $\boldsymbol{R}^{\text{d}}$ contains inter-disease relationships and has size $D^2$; $\boldsymbol{R}^{\text{g}}$ contains inter-group relationships and has size $K^2$. Consequently, the total number of relationship parameters is reduced to $D^2 + K^2 + 1$, significantly smaller than the original size $D^2 K^2$.

Moreover, Equation (4) shows that the original aggregation can be reformulated as a two-step process: first, parameters are separately aggregated using inter-disease and inter-group relationships to obtain two intermediate results; then, these results are aggregated to produce the final output. This makes the learning process easier and thus can bring about improved performance.

## 4.3 Third Design: A Novel Bayesian Network

### 4.3.1 Overview of the Third Design

With the first two designs, we reduce the complex "(disease, patient) - (disease, patient)" four-dimensional relationship to a scalar $a$ and two two-dimensional relationships $\boldsymbol{R}^{\text{d}}$ and $\boldsymbol{R}^{\text{g}}$. Since $\{a, \boldsymbol{R}^{\text{d}}, \boldsymbol{R}^{\text{g}}\}$ also need to be learned, we refer to them as *relationship parameters* to distinguish them from the parameters of the assessment model (short as *model parameters*).

There are dependencies among relationship parameters, model parameters, and model performance. At a higher level, model performance directly depends on model parameters, referred to as the *performance-model dependency*. Meanwhile, since model parameters are aggregated based on relationships, there is a direct dependence between model parameters and relationship parameters, referred to as the *relationship-model dependency*. Furthermore, since relationship parameters influence model parameters, which in turn influence model performance, there is also a dependence between model performance and relationship parameters, referred to as the *performance-relationship dependency*. Notably, this dependence disappears once the model parameters are given. At a more granular level, finer dependencies also exist. For example, as noted earlier in the discussion of the fourth challenge faced by BDH-MTL, the relationship parameters of a model's two components should be different yet mutually dependent.

Given the advantage of Bayesian networks in modeling complex dependencies, we propose a novel Bayesian network. Our Bayesian network not only captures the complex dependencies to address the third challenge of BDH-MTL, but also tackles the fourth challenge by creating two sets of dependent relationship parameters for the model's two components.

Our Bayesian network includes a generative process that models the dependencies by sequentially generating relationship parameters, model parameters, and model performance. It also includes an inference process that estimates parameters involved in the generative process. We will present the generative process in the next section, followed by a detailed description of the inference process in the subsequent section.

### 4.3.2 The Generative Process of Our Proposed Bayesian Network

The generative process includes three steps: 1) generating relationship parameters, including inter-disease relationship, inter-group relationship, and the relative weight, 2) generating model parameters conditioned on the relationship parameters, to capture relationship-model dependency, and 3) generating model performance conditioned on the model parameters and actual data, to capture the performance-model dependency and performance-relationship dependency. The generative process is illustrated in Figure 1, where shaded nodes represent observed variables and unshaded (white) nodes represent latent variables.

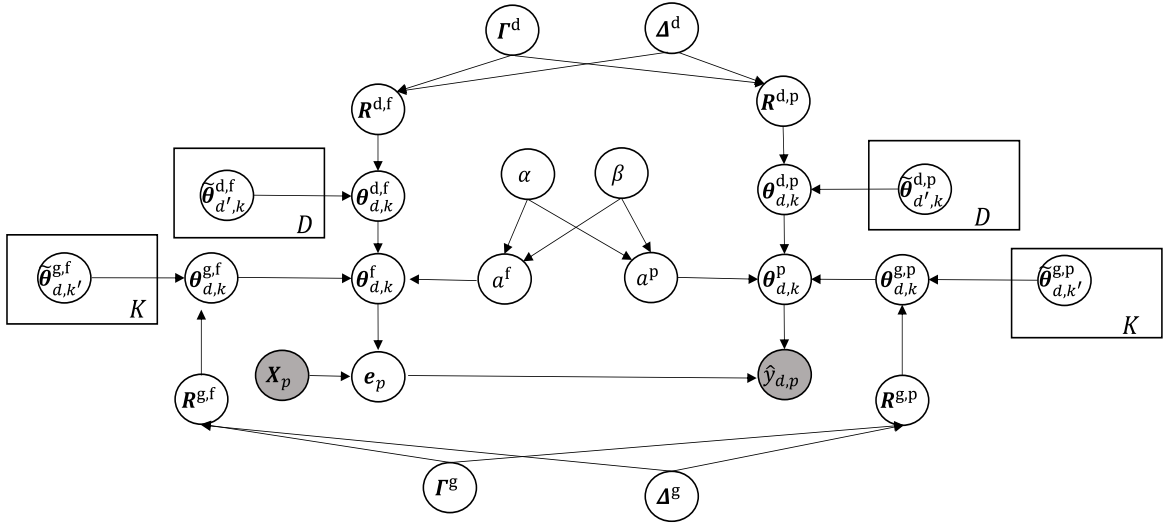

**Figure 1. The Generative Process of Our Bayesian Network**

Step 1: Generate Relationship Parameters

We introduce four matrices $\boldsymbol{R}^{d,f}$, $\boldsymbol{R}^{d,p}$, $\boldsymbol{R}^{g,f}$, and $\boldsymbol{R}^{g,p}$, for the inter-disease relationships of the feature extraction component, inter-disease relationships of the prediction component, inter-group relationships of the feature extraction component, and inter-group relationships of the prediction component, respectively. For instance, the element at $d$-th row and $d'$-th column of $\boldsymbol{R}^{d,f}$ (denoted as $R^{d,f}_{d,d'}$) represents the correlation between disease $d$ and disease $d'$ of the feature extraction component. Since $\boldsymbol{R}^{d,f}$ and $\boldsymbol{R}^{d,p}$ are separate, as are $\boldsymbol{R}^{g,f}$ and $\boldsymbol{R}^{g,p}$, we address the differences between the two components' relationships. Meanwhile, as we mentioned before, since $\boldsymbol{R}^{d,f}$ and $\boldsymbol{R}^{d,p}$ are for the two components of the same model, they should share some similarities. To incorporate this, we introduce a shared prior for them. Specifically, we adopt a matrix normal prior for each, i.e., $\boldsymbol{R}^{d,f} \sim \mathcal{MN}(\boldsymbol{0}, \boldsymbol{\Delta}^d, \boldsymbol{\Gamma}^d)$ and $\boldsymbol{R}^{d,p} \sim \mathcal{MN}(\boldsymbol{0}, \boldsymbol{\Delta}^d, \boldsymbol{\Gamma}^d)$, where $\mathcal{MN}$ represents matrix normal distribution. $\boldsymbol{\Delta}^d$ and $\boldsymbol{\Gamma}^d$ are two semipositive defined matrices with size $D \times D$. We share $\boldsymbol{\Delta}^d$ and $\boldsymbol{\Gamma}^d$ across $\boldsymbol{R}^{d,f}$ and $\boldsymbol{R}^{d,p}$ to impose a common prior to capture their similarities. Similarly, we assign matrix normal priors to $\boldsymbol{R}^{g,f}$ and $\boldsymbol{R}^{g,p}$: $\boldsymbol{R}^{g,f} \sim \mathcal{MN}(\boldsymbol{0}, \boldsymbol{\Delta}^g, \boldsymbol{\Gamma}^g)$, $\boldsymbol{R}^{g,p} \sim \mathcal{MN}(\boldsymbol{0}, \boldsymbol{\Delta}^g, \boldsymbol{\Gamma}^g)$.

Similarly, we introduce two separate relative weights $a^f$ and $a^p$ for each component. $a^f$ and $a^p$ are closely related and mutually dependent, and we assign them the same prior. Since both $a^f$ and $a^p$ ranges from 0 to 1, we adopt a Beta prior, i.e., $a^f \sim \text{Beta}(\alpha, \beta)$ and $a^p \sim \text{Beta}(\alpha, \beta)$. $\alpha$ and $\beta$ are hyperparameters.

To summarize, we generate two sets of relationship parameters: one consisting of $\boldsymbol{R}^{d,f}$, $\boldsymbol{R}^{g,f}$ and $a^f$ for the feature extraction component, and the other consisting of $\boldsymbol{R}^{d,p}$, $\boldsymbol{R}^{g,p}$ and $a^p$ for the prediction component. Meanwhile, by placing a shared prior on $\boldsymbol{R}^{d,f}$ and $\boldsymbol{R}^{d,p}$, a shared prior on $\boldsymbol{R}^{g,f}$ and $\boldsymbol{R}^{g,p}$, and a shared prior on $a^f$ and $a^p$, we capture both differences and similarities between the two components' relationship parameters. In this way, we address the fourth challenge of BDH-MTL. In addition, incorporating priors enhances learning stability and effectiveness by regularizing the model against unreliable parameters, a benefit widely evidenced by prior studies (Mohammad-Djafari 2021, Fortuin 2022).

Step 2: Generate Assessment Model Parameters

We denote the current parameters of the assessment models as $\{\widetilde{\boldsymbol{\theta}}_{d,k}^{d,f}|\forall d,k\}$ and $\{\widetilde{\boldsymbol{\theta}}_{d,k}^{g,f}|\forall d,k\}$. Based on the current parameters and the relationship parameters (i.e., $\boldsymbol{R}^{d,f}$, $\boldsymbol{R}^{g,f}$, $\boldsymbol{R}^{d,p}$, $\boldsymbol{R}^{g,p}$, $a^f$ and $a^p$), we then generate the model parameters for the feature extraction and the prediction components. Guided by Equation (5), each component's parameters are generated in two steps: first, intermediate parameters are obtained from the inter-disease and inter-group relationships; second, these intermediates are combined using the relative weight to produce the final parameters. We denote the parameter of the feature extraction component of model $M_{d,k}$ as $\boldsymbol{\theta}_{d,k}^{f}$ and that of the prediction component as $\boldsymbol{\theta}_{d,k}^{p}$. We use $\boldsymbol{\theta}_{d,k}^{f}$ as an example.

First, we use the relationships $\boldsymbol{R}^{d,f}$ and $\boldsymbol{R}^{g,f}$ to generate two intermediate parameters, $\boldsymbol{\theta}_{d,k}^{d,f}$ and $\boldsymbol{\theta}_{d,k}^{g,f}$:

$$\boldsymbol{\theta}_{d,k}^{d,f} = \frac{\sum_{d'=1}^{D} R_{d,d'}^{d,f} \cdot \widetilde{\boldsymbol{\theta}}_{d',k}^{d,f}}{\sum_{d'=1}^{D} R_{d,d'}^{d,f}}, \quad \boldsymbol{\theta}_{d,k}^{g,f} = \frac{\sum_{k'=1}^{K} R_{k,k'}^{g,f} \cdot \widetilde{\boldsymbol{\theta}}_{d,k'}^{g,f}}{\sum_{k'=1}^{K} R_{k,k'}^{g,f}} \quad (5)$$

Second, we generate the parameter $\boldsymbol{\theta}_{d,k}^{f}$ by combining $\boldsymbol{\theta}_{d,k}^{d,f}$ and $\boldsymbol{\theta}_{d,k}^{g,f}$ with the relative weight $a^f$:

$$\boldsymbol{\theta}_{d,k}^{f} = a^f \cdot \boldsymbol{\theta}_{d,k}^{d,f} + (1 - a^f) \cdot \boldsymbol{\theta}_{d,k}^{g,f} \quad (6)$$

Similarly, we generate the prediction component's parameter $\boldsymbol{\theta}_{d,k}^{p}$. In this way, we generate the parameters for the assessment model $M_{d,k}$.

Step 3: Generate Assessment Performance

With the parameters generated from Step 2, we can generate the assessment model's performance based on its predictions given the input data, which includes wearable sensor data and patient's profile information. Without loss of generality, assume a patient $p$ belongs to group $k$. Given patient $p$'s data $X_p$ (including $X_p^{\text{sensor}}$ and $X_p^{\text{profile}}$), the extracted feature is generated as $e_p = f_{(\theta_{d,k}^{\text{f}})}(X_p)$ where $f_{(\theta_{d,k}^{\text{f}})}$ is the function parameterized by $\theta_{d,k}^{\text{f}}$. The prediction is generated by $\hat{y}_{d,p} = f_{(\theta_{d,k}^{\text{p}})}(e_p)$ where $f_{(\theta_{d,k}^{\text{p}})}$ is the function parameterized by $\theta_{d,k}^{\text{p}}$ and $\hat{y}_{d,p}$ is the probability that the $d$-th disease is positive (i.e., $\hat{y}_{d,p} = 1$). The model's performance is measured by the likelihood of the ground truth, denoted as $p_{\text{BN}}(\hat{y}_{d,p}|X_p, \theta_{d,k}^{\text{f}}, \theta_{d,k}^{\text{p}})$, which we simplify to $p_{\text{BN}}(y_{d,p})$ to reduce notation clutter. If $y_{d,p} = 1$, $p_{\text{BN}}(y_{d,p})$ is exactly $\hat{y}_{d,p}$; if $y_{d,p} = 0$, $p_{\text{BN}}(y_{d,p})$ is exactly $1 - \hat{y}_{d,p}$. This way, we first generate the assessment model's predictions and then its performance.

By these three steps, our Bayesian network explicitly model the dependencies among relationship parameters (i.e., $R^{\text{d,f}}$, $R^{\text{g,f}}$, $R^{\text{d,p}}$, $R^{\text{g,p}}$, $a^{\text{f}}$ and $a^{\text{p}}$), the parameters of each assessment model (i.e., $\{\theta_{d,k}^{\text{f}}|\forall d, k\}$ and $\{\theta_{d,k}^{\text{p}}|\forall d, k\}$), and model performance (i.e., $\{p_{\text{BN}}(y_{d,p})|\forall d, p\}$), thus addressing the third challenge faced by BDH-MTL. Meanwhile, the Bayesian network also models the dependencies between the relationship parameters of a model's two components, thus addressing the fourth challenge.

### 4.3.3 The Inference Process of Our Proposed Bayesian Network

The generative process models dependencies, while the inference process estimates unknown parameters (including relationship and model parameters) by maximizing the likelihood of the observations, thereby enhancing the model's performance in disease assessment. We use the classic variational inference (Blei et al. 2017), first deriving an objective function and then optimizing it by updating the parameters.

(1) Objective Function

We denote the likelihood of the observations under our Bayesian network as $p_{\text{BN}}(X, Y)$, which can be factorized as $p_{\text{BN}}(X)p_{\text{BN}}(Y|X)$. Since $p_{\text{BN}}(X)$ is a constant with respect to our Bayesian network, maximizing $p_{\text{BN}}(X, Y)$ is equivalent to maximizing $p_{\text{BN}}(Y|X)$. Since each disease and group utilizes a separate assessment model, predictions are generated independently. Hence, $p_{BN}(Y|X)$ can be expanded as $\prod_{k=1}^{K} \prod_{G(p)=k} \prod_{d=1}^{D} p_{BN}(y_{d,p})$ where $G(p) = k$ refers to all patients belonging to group $k$. For stable learning, we maximize the log form of the likelihood, expressed as $\sum_{k=1}^{K} \sum_{G(p)=k} \sum_{d=1}^{D} \log p_{BN}(y_{d,p})$.

However, the computation of the log-likelihood requires us to infer the hidden variables in the generative process, including relationship parameters ($R^{d,f}$, $R^{g,f}$, $R^{d,p}$, $R^{g,p}$, $a^f$ and $a^p$) and model parameters ($\{\theta_{d,k}^f | \forall d,k\}$ and $\{\theta_{d,k}^p | \forall d,k\}$ denoted as $\Theta$).

For the relationship parameters, which are random variables with prior distributions, our goal is to infer their posterior distributions given the observed data. However, the true posterior is generally intractable. To address this, we adopt the classic variational inference algorithm. This algorithm introduces a tractable variational distribution to approximate the true posterior. The parameters of these random variables can then be obtained from this variational distribution. In our study, the variational distributions are designed as neural networks, referred to as inference networks, which enable them to flexibly approximate diverse posterior forms. We denote the variational distribution as $q_\psi(R^{d,f}, R^{g,f}, R^{d,p}, R^{g,p}, a^f, a^p)$. Then, we can derive a lower bound on the log-likelihood, known as the Evidence Lower Bound (ELBO) (details of the derivation are in Appendix E):

$$\log p_{BN}(Y|X) \geq \mathbb{E}_{q_\psi} \left[ \sum_{k=1}^{K} \sum_{G(p)=k} \sum_{d=1}^{D} \log p(y_{d,p} | R^{d,f}, R^{g,f}, R^{d,p}, R^{g,p}, a^f, a^p, \Theta, X_p) \right] \quad (7)$$
$$- \text{KL}((q_\psi(R^{d,f}, R^{g,f}, R^{d,p}, R^{g,p}, a^f, a^p) || p(R^{d,f}, R^{g,f}, R^{d,p}, R^{g,p}, a^f, a^p))$$

$p(y_{d,p} | R^{d,f}, R^{g,f}, R^{d,p}, R^{g,p}, a^f, a^p, \Theta, X_p)$ is the full expansion of $p_{BN}(y_{d,p})$. As mentioned earlier, for each disease $d$, if $y_{d,p} = 1$, the likelihood equals the predicted value $\hat{y}_{d,p}$ from the assessment model; if $y_{d,p} = 0$, it equals $1 - \hat{y}_{d,p}$. Hence, the log-likelihood for disease $d$ becomes $y_{d,p} \log \hat{y}_{d,p} + (1 - y_{d,p}) \log(1 - \hat{y}_{d,p})$, which is exactly the negative of the cross-entropy. Cross-entropy is a standard loss function in deep learning and has demonstrated excellent performance in classification tasks. Since maximizing the log-likelihood is equivalent to maximizing the negative cross-entropy, and thus to minimizing the cross-entropy, we actually also use the cross-entropy loss in our objective. Collectively, the first term of the ELBO (denoted as $\text{ELBO}_{\text{term1}}$) can be computed as:

$$\text{ELBO}_{\text{term1}} = \mathbb{E}_{q_\psi}\left[\sum_{k=1}^{K}\sum_{G(p)=k}\sum_{d=1}^{D} y_{d,p}\log \hat{y}_{d,p} + (1-y_{d,p})\log(1-\hat{y}_{d,p})\right] \quad (8)$$

For the second term, we follow the mean-field approximation (Tanaka 1998), assuming that the hidden variables ($\boldsymbol{R}^{d,f}, \boldsymbol{R}^{g,f}, \boldsymbol{R}^{d,p}, \boldsymbol{R}^{g,p}, a^f, a^p$) are independent. This simplifies the joint posterior inference into a set of lower-dimensional, tractable subproblems. The second term (denoted as $\text{ELBO}_{\text{term2}}$) becomes:

$$\text{ELBO}_{\text{term2}} = \text{KL}(q_\psi(\boldsymbol{R}^{d,f})||p(\boldsymbol{R}^{d,f})) + \text{KL}(q_\psi(\boldsymbol{R}^{g,f})||p(\boldsymbol{R}^{g,f})) + \text{KL}(q_\psi(\boldsymbol{R}^{d,p})||p(\boldsymbol{R}^{d,p}))$$
$$+ \text{KL}(q_\psi(\boldsymbol{R}^{g,p})||p(\boldsymbol{R}^{g,p})) + \text{KL}(q_\psi(a^f)||p(a^f)) + \text{KL}(q_\psi(a^p)||p(a^p)) \quad (9)$$

Here, $q_\psi(\boldsymbol{R}^{d,f})$ denotes the variational distribution, and $p(\boldsymbol{R}^{d,f})$ is the prior distribution (i.e., matrix normal distribution) of $\boldsymbol{R}^{d,f}$. Similar notations apply to the other variables. We construct an inference network for each variable, where each network takes the relevant observed variables as input and outputs the parameters of the corresponding variational distribution. The detailed structures of the inference networks are provided in Appendix F. Then, each KL term in Equation (9) can be computed in closed form. For instance, for $\text{KL}(q_\psi(a^f)||p(a^f))$, we model $q_\psi(a^f)$ as a Beta distribution, with the inference network producing parameters $\hat{\alpha}^f$ and $\hat{\beta}^f$ that define $\text{Beta}(\hat{\alpha}^f, \hat{\beta}^f)$. The KL divergence is then computed as:

$$\text{KL}(q_\psi(a^f)||p(a^f)) = \ln\left(\frac{\text{Beta}(\alpha,\beta)}{\text{Beta}(\hat{\alpha}^f,\hat{\beta}^f)}\right) + (\hat{\alpha}^f - \alpha)\psi(\hat{\alpha}^f) + (\hat{\beta}^f - \beta)\psi(\hat{\beta}^f) \\ -(\hat{\alpha}^f - \alpha + \hat{\beta}^f - \beta) \quad (10)$$

We compute $\text{KL}(q_\psi(a^p)||p(a^p))$ similarly. For $\text{KL}(q_\psi(\boldsymbol{R}^{d,f})||p(\boldsymbol{R}^{d,f}))$, we also define the $q_\psi(\boldsymbol{R}^{d,f})$ as the matrix normal distribution. However, the two-dimensional matrix form of $\boldsymbol{R}^{d,f}$ makes it challenging for the inference network to directly output such a matrix. To address this, we follow prior studies that transform the matrix normal distribution into an equivalent multivariate normal distribution using the Kronecker product (Viroli 2012). Specifically, a matrix normal distribution $\mathcal{MN}(\boldsymbol{0}, \boldsymbol{\Gamma}, \boldsymbol{\Delta})$ can be transformed into an equivalent multivariate normal distribution $\mathcal{N}(\text{vec}(\boldsymbol{0}), \boldsymbol{\Gamma} \otimes \boldsymbol{\Delta})$, where $\text{vec}(\boldsymbol{0})$ denotes the vectorized zero matrix, and $\otimes$ is the Kronecker product. The advantage of this transformation is that it

allows us to represent $\boldsymbol{R}^{\text{d,f}}$ in a one-dimensional vector form (i.e., as a sample from a multivariate normal distribution). We use inference networks to get $\hat{\boldsymbol{\mu}}^{\text{d}}$ and $\hat{\boldsymbol{\Delta}}^{\text{d}}$ and $\hat{\boldsymbol{\Gamma}}^{\text{d}}$ to define multivariate normal distribution $\mathcal{N}\big(\text{vec}(\hat{\boldsymbol{\mu}}^{\text{d}}), \hat{\boldsymbol{\Gamma}}^{\text{d}} \otimes \hat{\boldsymbol{\Delta}}^{\text{d}}\big)$, and equivalently the matrix normal distribution $\mathcal{MN}\big(\hat{\boldsymbol{\mu}}^{\text{d}}, \hat{\boldsymbol{\Delta}}^{\text{d}}, \hat{\boldsymbol{\Gamma}}^{\text{d}}\big)$. Then, we have:

$$\text{KL}(q_\psi(\boldsymbol{R}^{\text{d,f}}) || p(\boldsymbol{R}^{\text{d,f}})) = \frac{1}{2}\left[\text{vec}(-\hat{\boldsymbol{\mu}}^{\text{d}})^{\text{T}} \text{vec}\left((\hat{\boldsymbol{\Gamma}}^{\text{d}})^{-1}(-\hat{\boldsymbol{\mu}}^{\text{d}})(\hat{\boldsymbol{\Delta}}^{\text{d}})^{-1}\right)\right]$$
$$+ \frac{1}{2}\left[\text{tr}\left(\left((\hat{\boldsymbol{\Delta}}^{\text{d}})^{-1}\hat{\boldsymbol{\Delta}}^{\text{d}}\right) \otimes \left((\hat{\boldsymbol{\Gamma}}^{\text{d}})^{-1}\hat{\boldsymbol{\Gamma}}^{\text{d}}\right)\right) - K \ln \frac{|\hat{\boldsymbol{\Delta}}^{\text{d}}|}{|\boldsymbol{\Delta}^{\text{d}}|} - K \ln \frac{|\hat{\boldsymbol{\Gamma}}^{\text{d}}|}{|\boldsymbol{\Gamma}^{\text{d}}|} - K^2\right] \quad (11)$$

As is common practice, we set $\boldsymbol{\Delta}^{\text{d}}, \boldsymbol{\Gamma}^{\text{d}}$ as identity matrices. We compute other KL divergences similarly.

### 4.3.4 Maximizing the ELBO

With both the first and second terms computable, we can compute the ELBO. Since both the relationship parameters and the model parameters influence the ELBO, they must be updated to maximize it. To achieve this, we adopt a coordinate descent strategy and iteratively update the two types of parameters.

In Step 1, we first fix the relationship parameters $\boldsymbol{R}^{\text{d,f}}, \boldsymbol{R}^{\text{g,f}}, \boldsymbol{R}^{\text{d,p}}, \boldsymbol{R}^{\text{g,p}}, a^{\text{f}}, a^{\text{p}}$ while updating the model parameters $\boldsymbol{\Theta}$ to maximize the ELBO. The values of the relationship parameters are obtained by randomly sampling from their corresponding variational distributions. In this step, only the first term of the ELBO is relevant. Hence, this is equivalent to training the assessment models based on the current parameters $\widetilde{\boldsymbol{\Theta}}$ to minimize the cross-entropy loss on the dataset. Optimization algorithms such as Adam are employed, same as existing deep learning studies.

In Step 2, we fix the newly updated $\boldsymbol{\Theta}$ while update the relationship parameters $\boldsymbol{R}^{\text{d,f}}, \boldsymbol{R}^{\text{g,f}}, \boldsymbol{R}^{\text{d,p}}, \boldsymbol{R}^{\text{g,p}}, a^{\text{f}}, a^{\text{p}}$. In this step, both terms of the ELBO are involved. Similarly, optimization algorithms such as Adam are employed. These two steps are repeated until convergence, defined as when the change in ELBO between iterations falls below a set threshold. At convergence, we obtain the optimized $\boldsymbol{\Theta}$, containing parameters for all assessment models $M_{d,k}, \forall d, k$. In other words, a personalized assessment model is created for each disease $d$ and patient group $k$, thereby addressing the issue of double heterogeneity.

### 4.4 Methodological Novelty of This Study

Jointly modeling both disease and patient heterogeneity is essential for effective multi-disease assessment. However, existing MTL studies primarily focus on disease heterogeneity, with few considering patient heterogeneity at the same time. To address this gap, we propose a base method, BDH-MTL, that jointly considers disease and patient heterogeneity in multi-disease assessment. Although BDH-MTL primarily integrates existing techniques, it provides a foundation for understanding the challenges of addressing double heterogeneity. To address the challenges of BDH-MTL, we further propose ADH-MTL with three designs. Among them, one applies an existing method to our case (i.e., group-level modeling with K-means), while the other two are novel. Hence, one novelty lies in the decomposition method within ADH-MTL. Specifically, after group-level modeling, the resulting relationships form a four-dimensional matrix of the type "(disease, group) - (disease, group)." To manage this complexity, we propose a novel decomposition method that splits the four-dimensional matrix into two simpler two-dimensional matrices: inter-disease relationships and inter-group relationships. This significantly reduces complexity and effectively disentangles the two types of relationships, enabling more effective and efficient learning. The other methodological novelty is our Bayesian network to capture dependencies among model parameters, patient-disease relationships, and performance. Although a few studies have also proposed Bayesian networks to model dependencies in MTL, our Bayesian network differs in two key aspects. First, we consider both patient heterogeneity and disease heterogeneity, thus including two types of relationships (i.e., inter-disease and inter-group relationships). This not only increases the number of dependencies but also introduces a new type of dependency. Although the inter-disease and inter-group relationships are generated separately, they become dependent once the model parameters are produced. This dependency is captured by our Bayesian network. We further introduce relative weights, constrained between 0 and 1 and assigned to a Beta prior, to integrate these two relationships. Second, our study focuses on disease assessment using unstructured wearable data, and thus, we employ deep learning models. Since each deep learning model comprises two functionally distinct yet closely related components (i.e., the feature extraction component and the prediction component), we explicitly model the dependencies between them. Specifically, by placing shared priors on $R^{d,f}$ and $R^{d,p}$, on $R^{g,f}$ and $R^{g,p}$, and on $a^f$ and $a^p$, and then learning them via variational inference, we capture both the differences and similarities between the relationship parameters of the two components.

## 5. Empirical Analyses

### 5.1 Dataset and Evaluation Metrics

We utilize the National Health and Nutrition Examination Survey (NHANES) dataset, which provides labeled health outcomes for individuals. This study focuses on comorbid chronic diseases, including diabetes, cardiovascular disease, high cholesterol, and depression. These diseases affect gait patterns and can be detected through wearable sensor data collected in daily life (Pirker and Katzenschlager 2017). Depression is measured using the Patient Health Questionnaire (PHQ-9). Labels for diabetes, cardiovascular disease, and high cholesterol are derived from participants' responses to standardized health surveys. A full list of the survey items used to construct the labels is provided in Appendix G.1. The sensor data is collected from smartwatches. The dataset contains detailed profile information such as family disease history, overweight status, grip strength, and other related variables, which are in Appendix G.2. Information on data types and collection is in Appendix G.3. Our dataset consists of 1,785 patients, with 24% having diabetes, 26% cardiovascular disease, 55% high cholesterol, and 27% depression.

We use precision, recall, and F1 as the evaluation metrics. 80% of the patients in the dataset are allocated for training, while the remaining 20% are reserved for testing. We computed precision, recall, and F1 score for each disease. All experiments were repeated five times, with results reported as averages accompanied by standard deviations.

### 5.2 Experiments

#### 5.2.1 Experiment 1: Comparison with Single Disease Assessment Methods

We first examine whether jointly assessing multiple diseases improves performance compared to assessing each disease independently, where a model is independently trained for each disease. ADH-MTL represents the full version of our method. Hence, we compare our ADH-MTL with six baselines: SCNN (Uddin 2024), Stack LSTM (Nunavath et al. 2021), ResLSTM (Zhang et al. 2023), CNN-LSTM (Pallewar et al. 2024), LSTM-CNN (Zhao et al. 2023), CNN-LSTM with Attention (Deng et al. 2023), and STransformer (Dirgová Luptáková et al. 2022). The baselines were chosen to represent a diverse range of model architectures that have demonstrated effectiveness in handling wearable sensor data. Technical details of the baselines are included in Appendix H.

**Table 1 Comparison Between ADH-MTL and Single Disease Assessment Methods**

| Methods | Diabetes | | | Cardiovascular | | | Depression | | | High cholesterol | | |
|---|---|---|---|---|---|---|---|---|---|---|---|---|
| | F1 | Prec. | Rec. | F1 | Prec. | Rec. | F1 | Prec. | Rec. | F1 | Prec. | Rec. |
| SCNN | 0.6737 ±0.0174 | 0.6710 ±0.0470 | 0.7002 ±0.0221 | 0.6586 ±0.0132 | 0.6015 ±0.0197 | 0.7335 ±0.0640 | 0.6324 ±0.0472 | 0.6605 ±0.0636 | 0.6084 ±0.0409 | 0.6750 ±0.0138 | 0.5745 ±0.0144 | 0.8214 ±0.0119 |
| Stack LSTM | 0.6911 ±0.0164 | 0.6887 ±0.0203 | 0.6941 ±0.0092 | 0.6047 ±0.0197 | 0.5368 ±0.0338 | 0.6943 ±0.0080 | 0.6884 ±0.0353 | 0.6613 ±0.1625 | 0.7890 ±0.1548 | 0.6200 ±0.0354 | 0.5722 ±0.0242 | 0.6785 ±0.0615 |
| ResLSTM | 0.6058 ±0.0389 | 0.6107 ±0.0470 | 0.6243 ±0.1304 | 0.5760 ±0.0648 | 0.5543 ±0.0545 | 0.7006 ±0.1912 | 0.7632 ±0.0569 | 0.6846 ±0.0578 | 0.8625 ±0.0534 | 0.6205 ±0.0734 | 0.5575 ±0.0674 | 0.7143 ±0.1331 |
| CNN-LSTM | 0.6744 ±0.0038 | 0.6663 ±0.0353 | 0.6858 ±0.0283 | 0.6199 ±0.0636 | 0.5929 ±0.0390 | 0.6813 ±0.1641 | 0.7440 ±0.0424 | 0.7437 ±0.0540 | 0.7727 ±0.1607 | 0.6758 ±0.0285 | 0.5720 ±0.0456 | 0.8302 ±0.0268 |
| LSTM-CNN | 0.6655 ±0.0152 | 0.6619 ±0.0199 | 0.6723 ±0.0453 | 0.6995 ±0.0188 | 0.6154 ±0.0313 | 0.7571 ±0.0706 | 0.7027 ±0.0328 | 0.6833 ±0.0198 | 0.7225 ±0.0750 | 0.6878 ±0.0174 | 0.6024 ±0.0345 | 0.8054 ±0.0339 |
| CNN-LSTM with Attention | 0.6737 ±0.0074 | 0.6408 ±0.0503 | 0.7196 ±0.0591 | 0.6525 ±0.0337 | 0.5963 ±0.0337 | 0.7216 ±0.0431 | 0.7588 ±0.0463 | 0.7352 ±0.0629 | 0.7849 ±0.0281 | 0.6728 ±0.0058 | 0.5892 ±0.0307 | 0.7894 ±0.0447 |
| STransformer | 0.6712 ±0.0196 | 0.6626 ±0.0451 | 0.6838 ±0.0342 | 0.6646 ±0.0299 | 0.5697 ±0.0438 | 0.8013 ±0.0278 | 0.6967 ±0.0385 | 0.7049 ±0.0603 | 0.7033 ±0.0967 | 0.6637 ±0.0243 | 0.5767 ±0.0234 | 0.7875 ±0.0747 |
| ADH-MTL | **0.7420 ±0.0005** | **0.7029 ±0.0224** | **0.7876 ±0.0290** | **0.7187 ±0.0063** | **0.6240 ±0.0260** | **0.8543 ±0.0664** | **0.8716 ±0.0337** | **0.7868 ±0.0403** | **0.9775 ±0.0226** | **0.7272 ±0.0228** | **0.6579 ±0.0805** | **0.8326 ±0.0707** |

As shown in Table 1, our ADH-MTL consistently outperforms single disease assessment baselines. Taking depression as an example, ADH-MTL attains an F1 score of 0.8716, representing improvements of 14.87% and 17.15% over the suboptimal baselines CNN-LSTM with Attention and CNN-LSTM, respectively. The advantage of ADH-MTL arises from two key factors. First, joint assessment leverages shared information across diseases while retaining disease-specific features. Second, the personalized model accounts for double heterogeneity, i.e., the heterogeneity across both patients and diseases. Overall, the results demonstrate that our ADH-MTL outperforms methods that assess each disease independently.

### 5.2.2 Experiment 2: Comparison with Multi-Disease Assessment Methods

Experiment 2 compares our ADH-MTL with existing MTL baselines. Based on the literature review in Section 2, two categories of baselines are included. The first category, feature-sharing methods, includes four baselines: AMM (Zhang et al. 2021), MAInt (Yin et al. 2024), BilzNet (Tan et al. 2023), and DynaShare (Rahimian et al. 2023), respectively representing the methods of weighted aggregation of shared features, feature selection from shared features, simultaneously learning task-specific features, and dynamic module selection. The second category, parameter-sharing methods, consists of distribution-based, matrix decomposition-based, and relationship-based methods. As noted above, the first two types mainly apply to classical machine learning models (e.g., SVM) with vector-based parameters. Since our model is a deep learning model, we focus on the third type, including similarity-regularization-based methods such as JTR (Nishi et al. 2024) and CRL (Sang et al. 2024), and parameter-aggregation-based methods such as Fedbone (Chen et al. 2024) and MAO (Hervella et al. 2024). The core ideas of these baselines are summarized in the literature review (Section 2), while their technical details are provided in Appendix I.

Table 2 Comparison Between ADH-MTL and Multi-Disease Assessment Methods

| Methods | Diabetes | | | Cardiovascular | | | Depression | | | High cholesterol | | |
|---|---|---|---|---|---|---|---|---|---|---|---|---|
| | F1 | Prec. | Rec. | F1 | Prec. | Rec. | F1 | Prec. | Rec. | F1 | Prec. | Rec. |
| AMM | 0.6428 ±0.0388 | 0.6767 ±0.0928 | 0.6729 ±0.2085 | 0.6032 ±0.0381 | 0.5460 ±0.0401 | 0.7188 ±0.1988 | 0.6193 ±0.1216 | 0.7056 ±0.2947 | 0.7131 ±0.1963 | 0.5258 ±0.0962 | 0.5735 ±0.0558 | 0.5769 ±0.2992 |
| MAInt | 0.5462 ±0.0540 | 0.5731 ±0.1262 | 0.5409 ±0.0671 | 0.5077 ±0.0793 | 0.5352 ±0.0531 | 0.5528 ±0.2501 | 0.5234 ±0.0224 | 0.4053 ±0.1049 | 0.6170 ±0.1951 | 0.5918 ±0.0901 | 0.5627 ±0.0351 | 0.6884 ±0.1710 |
| BlitzNet | 0.6426 ±0.0618 | 0.5832 ±0.0532 | 0.7156 ±0.0732 | 0.5122 ±0.0710 | 0.5373 ±0.0753 | 0.4870 ±0.1180 | 0.5014 ±0.0671 | 0.4961 ±0.1176 | 0.5194 ±0.0100 | 0.6001 ±0.0193 | 0.5571 ±0.0601 | 0.6852 ±0.1366 |
| DynaShare | 0.3337 ±0.0617 | 0.4794 ±0.2109 | 0.3311 ±0.1305 | 0.3693 ±0.2215 | 0.5121 ±0.0050 | 0.4292 ±0.4065 | 0.3096 ±0.1348 | 0.4015 ±0.1122 | 0.4519 ±0.3969 | 0.5995 ±0.0437 | 0.5669 ±0.1057 | 0.7040 ±0.2100 |
| JTR | 0.4512 ±0.1388 | 0.6173 ±0.1495 | 0.3574 ±0.1221 | 0.6644 ±0.0742 | **0.7251** **±0.1156** | 0.6046 ±0.1521 | 0.6618 ±0.0577 | 0.8116 ±0.1263 | 0.5628 ±0.0284 | 0.4992 ±0.0566 | 0.6009 ±0.0203 | 0.4355 ±0.0910 |
| CRL | 0.5263 ±0.1287 | 0.6411 ±0.1229 | 0.4527 ±0.1287 | 0.6098 ±0.0783 | 0.5841 ±0.0489 | 0.6494 ±0.1313 | 0.5363 ±0.1206 | 0.5418 ±0.0383 | 0.5735 ±0.2392 | 0.5363 ±0.0800 | 0.6388 ±0.1429 | 0.4693 ±0.0589 |
| Fedbone | 0.6729 ±0.0772 | 0.6934 ±0.1101 | 0.6559 ±0.0473 | 0.6976 ±0.0310 | 0.6783 ±0.0560 | 0.7204 ±0.0026 | 0.6732 ±0.1141 | **0.8182** **±0.1429** | 0.5719 ±0.0948 | 0.5897 ±0.1046 | 0.6356 ±0.1526 | 0.5537 ±0.0666 |
| MAO | 0.6698 ±0.0135 | 0.6807 ±0.0842 | 0.6833 ±0.1045 | 0.6422 ±0.0747 | 0.5240 ±0.0524 | 0.8308 ±0.1205 | 0.5288 ±0.0775 | 0.6418 ±0.0548 | 0.4789 ±0.1471 | 0.6568 ±0.0651 | 0.6458 ±0.1221 | 0.7053 ±0.1607 |
| ADH-MTL | **0.7420** **±0.0005** | **0.7029** **±0.0224** | **0.7876** **±0.0290** | **0.7187** **±0.0063** | 0.6240 ±0.0260 | **0.8543** **±0.0664** | **0.8716** **±0.0337** | 0.7868 ±0.0403 | **0.9775** **±0.0226** | **0.7272** **±0.0228** | 0.6579 ±0.0805 | **0.8326** **±0.0707** |

As shown in Table 2, the ADH-MTL consistently outperforms all MTL baselines. For diabetes, it achieves an F1 score of 0.7420, exceeding the best-performing baseline (Fedbone, F1 = 0.6729) by approximately 10.27%. For depression, ADH-MTL achieves an F1 score of 0.8716, representing a substantial 14.3% improvement over the Fedbone method (F1 = 0.7270). In the case of cardiovascular disease, ADH-MTL achieves the highest F1 score (0.7187), outperforming Fedbone (F1 = 0.6732). Although Fedbone demonstrates a high precision (0.8182), its low recall (0.5719) suggests limited coverage of positive cases, undermining its overall utility. Finally, for high cholesterol, ADH-MTL achieves an F1 score of 0.7272, significantly outperforming all baselines and demonstrating its superior capability. The advantage of ADH-MTL is that, unlike the baselines, it accounts not only for task heterogeneity but also for patient heterogeneity, enabling more personalized disease assessments.

### 5.2.3 Experiment 3: Ablation Analysis

We conducted an ablation study to evaluate the effectiveness of our key design choices. First, to demonstrate the advantages of the improvements in ADH-MTL, we compared it with BDH-MTL. Since its individual-level modeling is unsuitable for new patients, we also included the group-level modeling in it. Second, since ADH-MTL jointly models double heterogeneity, we constructed two additional variants: one removing disease heterogeneity by applying the same model across all diseases (short as w/o DH), and the other removing patient heterogeneity by applying the same model across all patients (short as w/o PH). Third, to examine the benefits of modeling distinct relationships between the feature extraction and prediction components, we tested a variant in which both components share the same relationship parameters. This variant is denoted as "w/o DR." The corresponding results are summarized in Table 3.

Table 3 Results of the Ablation Analysis

| Methods | Diabetes | | | Cardiovascular | | | Depression | | | High cholesterol | | |
|---|---|---|---|---|---|---|---|---|---|---|---|---|
| | F1 | Prec. | Rec. | F1 | Prec. | Rec. | F1 | Prec. | Rec. | F1 | Prec. | Rec. |
| BDH-MTL | 0.7341 ±0.0218 | 0.6537 ±0.0096 | **0.8396 ±0.0591** | 0.5991 ±0.0109 | 0.5547 ±0.0548 | 0.6435 ±0.0765 | 0.6126 ±0.0340 | 0.6224 ±0.0557 | 0.6586 ±0.1302 | 0.5893 ±0.0169 | 0.5163 ±0.0270 | 0.6878 ±0.0149 |
| w/o DH | 0.6985 ±0.0132 | 0.6445 ±0.0327 | 0.7659 ±0.0349 | 0.6565 ±0.0265 | 0.5831 ±0.0343 | 0.7545 ±0.0490 | 0.6940 ±0.0399 | 0.6472 ±0.0502 | 0.7832 ±0.1681 | 0.6232 ±0.0273 | 0.5811 ±0.0374 | 0.6768 ±0.0581 |
| w/o PH | 0.6939 ±0.0238 | 0.6270 ±0.0252 | 0.7768 ±0.0210 | 0.6460 ±0.0361 | 0.6055 ±0.0322 | 0.6924 ±0.0409 | 0.7445 ±0.0215 | 0.5934 ±0.0272 | **0.9890 ±0.0102** | 0.6439 ±0.0358 | 0.5943 ±0.0174 | 0.7038 ±0.0609 |
| w/o DR | 0.6649 ±0.0052 | 0.6299 ±0.0254 | 0.7543 ±0.0362 | 0.6710 ±0.0400 | 0.5838 ±0.0383 | 0.7908 ±0.0568 | 0.7119 ±0.0408 | 0.6887 ±0.0465 | 0.7371 ±0.0365 | 0.6363 ±0.0207 | 0.5531 ±0.0105 | 0.7495 ±0.0386 |
| ADH-MTL | **0.7420 ±0.0005** | **0.7029 ±0.0224** | 0.7876 ±0.0290 | **0.7187 ±0.0063** | **0.6240 ±0.0260** | **0.8543 ±0.0664** | **0.8716 ±0.0337** | **0.7868 ±0.0403** | 0.9775 ±0.0226 | **0.7272 ±0.0228** | **0.6579 ±0.0805** | **0.8326 ±0.0707** |

The results reveal four key findings. First, the performance of BDH-MTL is inferior to that of ADH-MTL. Although BDH-MTL performs relatively well on diabetes, it performs poorly on cardiovascular disease, depression, and high cholesterol. For instance, on cardiovascular disease, the F1 score decreases from 0.7187 to 0.5991. This suggests that merely integrating existing techniques to address double heterogeneity is limited, and novel designs in ADH-MTL are needed for better performance. Second, BDH-MTL performs comparably to strong baselines such as Fedbone and CRL in Experiment 2, showing that considering double heterogeneity is beneficial and that a straightforward approach can match baseline performance. Thus, BDH-MTL lays the foundation for addressing double heterogeneity. Third, removing either the disease or patient heterogeneity design reduces performance. For example, without considering disease heterogeneity (w/o DH), the F1 score for depression decreases from 0.8716 to 0.6940. Similarly, without considering patient heterogeneity (w/o PH), the F1 score for cardiovascular disease falls from 0.7187 to 0.6460. These results demonstrate the effectiveness of simultaneously addressing both heterogeneities. Fourth, sharing relationship parameters for the feature extraction and the prediction component leads to performance degradation. For instance, on diabetes, the F1 score drops from 0.7420 to 0.6649. This highlights the necessity of modeling the distinct relationships between the two components.

**5.2.4 Experiment 4: The Generalizability of Our Method on Different Patient Populations**

To evaluate the generalizability of our method across different patient populations, we stratify patients based on four demographic dimensions: age, income, race, and gender. For age, we use the median age in our dataset to divide patients into those older than 70 years (denoted as Senior) and those aged 70 or younger (denoted as Younger). For income, we apply a similar approach, categorizing patients into high-income and low-income groups based on the median annual income in our dataset. Regarding race, given that white individuals constitute the majority in the dataset, we assign them to one group and all other racial backgrounds to the second group. For gender, patients are categorized into male and female groups. We compare our ADH-MTL with a competing baseline (Fedbone) by examining the gap between two groups, where a smaller gap indicates better generalizability. The results are reported in Table 4.

Table 4 Disease Assessment Results in Different Patient Populations

| | | | Diabetes | | | Cardiovascular | | | Depression | | | High cholesterol | | |
|---|---|---|---|---|---|---|---|---|---|---|---|---|---|---|
| | | Group | F1 | Prec. | Rec. | F1 | Prec. | Rec. | F1 | Prec. | Rec. | F1 | Prec. | Rec. |
| Age | ADH-MTL | Younger | 0.8323 | 0.7804 | 0.8945 | 0.7348 | 0.6519 | 0.8436 | 0.8343 | 0.7224 | 0.9954 | 0.7495 | 0.7063 | 0.8026 |
| | | Senior | 0.7546 | 0.7353 | 0.8061 | 0.7204 | 0.6747 | 0.8088 | 0.8773 | 0.7815 | 0.9112 | 0.7039 | 0.6420 | 0.7888 |
| | | Gap | **0.0777** | **0.0451** | **0.0884** | **0.0144** | **0.0228** | 0.0348 | **0.0430** | **0.0591** | 0.0842 | **0.0456** | **0.0643** | **0.0138** |
| | Fedbone | Younger | 0.3586 | 0.3952 | 0.3824 | 0.6299 | 0.7622 | 0.5673 | 0.4897 | 1.0000 | 0.3298 | 0.5110 | 0.4910 | 0.5970 |
| | | Senior | 0.5051 | 0.6697 | 0.4068 | 0.5962 | 0.7019 | 0.5392 | 0.4323 | 0.7405 | 0.3069 | 0.4563 | 0.5573 | 0.4479 |
| | | Gap | 0.1465 | 0.2745 | **0.0244** | 0.0337 | 0.0603 | **0.0281** | 0.0574 | 0.2595 | **0.0229** | 0.0547 | 0.0663 | 0.1491 |
| Income | ADH-MTL | Low | 0.8157 | 0.7475 | 0.8987 | 0.7646 | 0.6765 | 0.8872 | 0.8184 | 0.7305 | 0.9333 | 0.8016 | 0.7217 | 0.9137 |
| | | High | 0.8732 | 0.8357 | 0.9254 | 0.733 | 0.6507 | 0.8481 | 0.8831 | 0.792 | 0.9323 | 0.7019 | 0.5841 | 0.8824 |
| | | Gap | **0.0575** | **0.0882** | **0.0267** | **0.0316** | **0.0258** | 0.0391 | **0.0647** | **0.0615** | **0.0010** | 0.0997 | **0.1376** | **0.0313** |
| | Fedbone | Low | 0.6262 | 0.6362 | 0.6584 | 0.6509 | 0.7848 | 0.5808 | 0.6426 | 0.9534 | 0.5098 | 0.5496 | 0.4585 | 0.6876 |
| | | High | 0.3454 | 0.3349 | 0.4370 | 0.5769 | 0.6347 | 0.5483 | 0.6475 | 0.9380 | 0.4984 | 0.6039 | 0.6659 | 0.6115 |

| | | | | | | | | | | | | | |
|---|---|---|---|---|---|---|---|---|---|---|---|---|---|
| RACE | ADH-MTL | Gap | 0.2808 | 0.3013 | 0.2214 | 0.0740 | 0.1501 | **0.0325** | **0.0049** | **0.0154** | 0.0114 | **0.0543** | 0.2074 | 0.0761 |
| | | White | 0.8053 | 0.7794 | 0.8364 | 0.7089 | 0.5867 | 0.8963 | 0.8856 | 0.795 | 0.9545 | 0.7241 | 0.649 | 0.8203 |
| | | Others | 0.7475 | 0.6606 | 0.8622 | 0.7619 | 0.6922 | 0.8569 | 0.8781 | 0.7837 | 1 | 0.7858 | 0.7302 | 0.8656 |
| | | Gap | **0.0578** | **0.1188** | **0.0258** | **0.0530** | **0.1055** | **0.0394** | **0.0075** | **0.0113** | **0.0455** | **0.0617** | **0.0812** | **0.0453** |
| | Fedbone | White | 0.5014 | 0.5331 | 0.4950 | 0.7048 | 0.8238 | 0.6181 | 0.5000 | 0.9257 | 0.3540 | 0.3630 | 0.4709 | 0.3541 |
| | | Others | 0.4026 | 0.6049 | 0.3799 | 0.6726 | 0.6137 | 0.7651 | 0.5499 | 0.9450 | 0.4013 | 0.6200 | 0.6196 | 0.6434 |
| | | Gap | 0.0988 | **0.0718** | 0.1151 | **0.0322** | 0.2101 | 0.1470 | 0.0499 | 0.0193 | 0.0473 | 0.2570 | 0.1487 | 0.2893 |
| Gender | ADH-MTL | Male | 0.8213 | 0.7795 | 0.8716 | 0.717 | 0.7008 | 0.7558 | 0.9436 | 0.8941 | 1 | 0.732 | 0.7315 | 0.7487 |
| | | Female | 0.7799 | 0.7179 | 0.8713 | 0.7202 | 0.6629 | 0.8402 | 0.8415 | 0.7289 | 0.9015 | 0.75 | 0.655 | 0.8809 |
| | | Gap | **0.0414** | **0.0616** | **0.0003** | **0.0032** | **0.0379** | **0.0844** | 0.1021 | **0.1652** | **0.0985** | **0.0180** | **0.0765** | **0.1322** |
| | Fedbone | Male | 0.4000 | 0.4545 | 0.3571 | 0.6288 | 0.5576 | 0.7252 | 0.5143 | 0.9039 | 0.3618 | 0.4687 | 0.5321 | 0.4781 |
| | | Female | 0.6728 | 0.5908 | 0.8531 | 0.6366 | 0.8382 | 0.5458 | 0.4153 | 0.7642 | 0.2853 | 0.5215 | 0.4089 | 0.7457 |
| | | Gap | 0.2728 | 0.1363 | 0.4960 | 0.0078 | 0.2806 | 0.1794 | **0.0990** | 0.1397 | 0.0765 | 0.0528 | 0.1232 | 0.2676 |

As shown in Table 4, ADH-MTL consistently achieves strong generalizability across all patient populations. For example, in assessing the severity of diabetes, the F1 score gap in ADH-MTL is 0.0777 between younger and senior groups, 0.0575 between low- and high-income groups, 0.0578 between White and other racial groups, and 0.0414 between male and female groups. By comparison, the Fedbone exhibits larger F1 score gaps: 0.1465 between younger and senior groups, 0.2808 between low- and high-income groups, 0.0988 between White and other racial groups, and 0.2728 between male and female groups. These results underscore our method's superior generalizability across diverse patient populations. This advantage may stem from our consideration of patient heterogeneity, allowing the development of personalized models tailored to different patient groups. In contrast, existing methods such as Fedbone adopt population-level modeling, which may favor the majority and fail to generalize to the minority.

### 5.3 Case Study and Practical Application Value

Since the main advantage of our ADH-MTL over the baselines lies in effectively modeling patient heterogeneity, we present a case study to demonstrate this capability. We selected four patients from different groups, including two pairs. The two patients within each pair have wearable sensor data of over 94% similarity. Following Ferrari et al. (2023), this similarity was measured by first extracting statistical features such as mean and variance from 15-minute intervals of wearable sensor data. These features were then reduced via Principal Component Analysis (PCA), and the resulting representations were used to compute cosine similarity. The main difference between the two patients within a pair lies in their profile information. We applied our ADH-MTL and Fedbone, the top-performing baseline, to predict disease presence. Our goal is to show that, even when sensor data input is similar, our model is able to more accurately predict diseases, due to our model's ability to discern patient heterogeneity. The results are shown in Table 5, with incorrect predictions highlighted in bold.

**Table 5 Comparison of Prediction Results on Patients with Similar Wearable Sensor Data**

| Patient ID | Sensor data similarity | Profile information | Ground-truth | ADH-MTL's predictions | Fedbone's predictions |
|---|---|---|---|---|---|
| Patient 76490 (Group 0) | 0.9516 | Asthma family history: N<br>Grip strength: Stronger<br>Overweight status: N<br>Physical inactivity: N | Diabetes: N | N (√) | N (√) |
| | | | Cardiovascular: N | N (√) | N (√) |
| | | | Depression: N | N (√) | N (√) |
| | | | High cholesterol: N | N (√) | N (√) |
| Patient 83006 (Group 2) | | Asthma family history: Y<br>Grip strength：Weaker<br>Overweight status: Y<br>Physical inactivity: Y | Diabetes: N | N (√) | N (√) |
| | | | Cardiovascular: N | N (√) | N (√) |
| | | | Depression: Y | Y (√) | **Y (✕)** |
| | | | High cholesterol: Y | Y (√) | **Y (✕)** |
| Patient 73567 (Group 1) | 0.9491 | Heart disease family history: Y<br>Physical inactivity: N<br>Blood pressure: Higher | Diabetes: Y | Y (√) | **Y (✕)** |
| | | | Cardiovascular: N | N (√) | N (√) |
| | | | Depression: N | N (√) | N (√) |
| | | | High cholesterol: N | N (√) | N (√) |
| Patient 77992 (Group 3) | | Heart disease family history: N<br>Physical inactivity: Y<br>Blood pressure: Lower | Diabetes: N | N (√) | N (√) |
| | | | Cardiovascular: N | N (√) | N (√) |
| | | | Depression: N | N (√) | N (√) |
| | | | High cholesterol: N | N (√) | N (√) |

Note: "Y" : the presence of the feature or disease; "N": absence; " √": correct prediction, "✕" : incorrect prediction.

Two key insights can be drawn. First, although patients may display highly similar behavioral patterns in sensor data, their disease statuses can differ markedly, highlighting patient heterogeneity. For instance, Patients 76490 and 83006 share a behavioral similarity of 0.9516; however, Patient 76490 remains healthy across all four diseases, whereas Patient 83006 has depression and high cholesterol. These discrepancies can be attributed to variations in their profiles: Patient 76490 demonstrates stronger grip strength, whereas Patient 83006 has a family history of asthma, is overweight, and exhibits physical inactivity.

Second, compared with Fedbone, our ADH-MTL more effectively captures patient heterogeneity. Specifically, although Patients 76490 and 83006 exhibit highly similar behavioral patterns, our method distinguishes their profile differences and assigns them to different groups (Group 0 and Group 2). Consequently, distinct assessment models are applied to the two patients. As a result, even with similar input sensor data, the predicted outcomes differ: our method correctly identifies that Patient 83006 suffers from both depression and high cholesterol, whereas Patient 76490 does not. In contrast, Fedbone produces identical predictions for both patients, indicating neither has depression nor high cholesterol, which is incorrect for Patient 83006. Similar results are observed for Patients 73567 and 77992, who show highly similar behaviors but differ in disease status. Our method correctly predicts Patient 73567 as diabetic and Patient 77992 as healthy, whereas Fedbone fails to capture this distinction, giving incorrect predictions for Patient 73567. These results are expected, as the baseline method applies the same set of assessment models to all patients, inevitably yielding identical predictions for highly similar inputs. In contrast, our method explicitly models patient heterogeneity, allowing for more accurate differentiation among patients with similar behavioral patterns but distinct clinical conditions.

To further evaluate the practical applicability of our ADH-MTL in chronic disease management, we conducted unstructured interviews with four physicians from a top hospital in China. The expert panel included three physicians specializing in chronic disease treatment, each with more than eight years of clinical experience, and one psychiatrist with fifteen years of clinical practice. We interviewed each expert for 30 minutes. Specifically, we introduced our method and its prediction results, including its overall evaluation performance and its ability to accurately distinguish patients who exhibit similar behavioral patterns but differ in their disease status, to the experts, followed by four guiding questions: 1) Do you think our method has practical application value in the continuous monitoring of patients' diseases? 2) Do you think our method can facilitate the timely screening and assessment of patients' health conditions and help optimize subsequent clinical treatment plans? 3) Would you be willing to use the assessment results generated by our method in your clinical practice? 4) Do you have any other comments regarding our method or results? The first three questions were measured using a five-point Likert scale (1 = Not at all, 2 = Somewhat, 3 = Neutral, 4 = Moderately, 5 = Very Much). After completing the structured questions, experts were invited to freely share their thoughts and suggestions based on their professional experience. The average ratings for the first three questions were 4.25, 4.50, and 4.50 out of 5.00, respectively, suggesting highly positive perceptions of our method's practical utility. For Question 4, we applied inductive thematic analysis to the interview transcripts (Thomas, 2006). The results were synthesized into three overarching themes, each comprising several specific subthemes. A summary of these findings is presented in Table 6.

Table 6. Summary of Inductive Analysis

| Themes | Sub-themes |
|---|---|
| Objectivity and Clinical Potential of Wearable Devices | **Objectivity and Behavioral Monitoring**<br>- Wearable devices provide more objective data compared to patients' self-reports.<br>- A core symptom of depression is reduced energy and vitality, which wearable devices can effectively capture, while traditional Q&A cannot.<br>- The objectivity of wearable devices benefits long-term medical development.<br>**Supporting Clinical Screening and Diagnosis**<br>- The proposed method can assist hospitals in preliminary screening and improve efficiency.<br>- It can support clinicians in auxiliary diagnosis. |
| Clinical Implications of Collaborative Assessment and Timely Intervention | **Necessity of Integrated Assessment**<br>- Collaborative assessment is more effective. Chronic disease patients are more likely to experience depression, and depression may affect metabolism, hindering recovery from other diseases.<br>**Importance of Timely Intervention**<br>- Timely intervention can promote patient recovery, especially when there are noticeable changes in the patient's condition. Such interventions are particularly beneficial. |
| Limitations of Traditional Self-Reported Assessments | **Subjectivity and Expression Difficulties**<br>- Patients tend to be subjective, and their judgments are not always accurate.<br>- Some patients cannot clearly express their feelings.<br>- Many patients are evasive or passively participate in surveys and are unwilling to express their real problems.<br>**Emotional and Symptomatic Fluctuations**<br>- Patients' disease states tend to fluctuate over time. For example, some depressed patients are active in the morning but become withdrawn in the evening.<br>- Emotional fluctuations make clinicians rely on experience-based judgment, lacking objective information for support.<br>**Individual Differences among Patients**<br>- Symptoms vary significantly across patients; each individual presents different manifestations. |

The interview offers valuable insights into how wearable sensor–based chronic disease assessment can enhance clinical practice. A prominent theme that emerged concerns the limitations of traditional self-reported assessments. Patients often exhibit subjective bias, have difficulty articulating their experiences, and may show emotional fluctuations that hinder accurate clinical judgment. In contrast, experts highlighted the objectivity and consistency of wearable devices, emphasizing their ability to capture subtle behavioral changes, such as reduced vitality in depressed patients that traditional questionnaires often fail to detect. Moreover, experts noted that such data-driven assessments could assist hospitals in preliminary screening and provide valuable support for diagnostic decision-making. Another key point concerns the potential for collaborative assessment and timely intervention. Chronic disease patients are often comorbid with depression, and continuous monitoring enables earlier and more targeted responses. Overall, these discussions suggest that wearable sensor–based assessments not only complement existing diagnostic methods but also pave the way for proactive, data-informed, and patient-centered healthcare practices.

## 6. Discussion and Conclusion

Assessing comorbid chronic diseases and depression is critical for collaborative care and integrated health management. Positioned as computational design research in IS, we propose BDH-MTL and ADH-MTL. BDH-MTL straightforwardly integrates existing techniques, providing a foundation for addressing double heterogeneity. ADH-MTL further addresses BDH-MTL's challenges through novel designs. We conduct comprehensive experiments on a real-world dataset, demonstrating the superiority of our method over baselines. Case studies further demonstrate our method's effectiveness and clinical relevance.

### 6.1 Contributions to IS

Our work is situated within the design science research (DSR) in IS, which emphasizes the development of computational artifacts to address business and societal challenges while contributing methodologically (Padmanabhan et al., 2022). Aligned with the pathways framework for design research (Abbasi et al., 2024), we propose a novel predictive method named ADH-MTL. The development of ADH-MTL is motivated by a pressing domain need: to jointly model comorbid chronic diseases and depression. This integrated approach allows the interrelationships among diseases to be exploited, thereby improving the assessment of each disease. Moreover, incorporating depression provides patients, families, and clinicians with a more holistic understanding of health status, supporting more appropriate interventions. Although numerous MTL methods have been proposed, most focus primarily on disease heterogeneity without adequately accounting for patient heterogeneity. Our study introduces a novel MTL method that effectively incorporates both disease heterogeneity and patient heterogeneity, demonstrating richer domain adaptation for personalized collaborative care.

Meanwhile, our study contributes to the cumulative tradition of IS research, particularly within the emerging domain of wearable sensor-based chronic disease management. While numerous studies in top-tier IS journals have made valuable advances, they have predominantly focused on individual diseases such as Parkinson's disease, fall risk, or depression. This study advances the state of the art by enabling the simultaneous management of multiple comorbid chronic diseases and depression, thereby facilitating collaborative care and integrated health management.

Considering the double heterogeneity in multi-task learning presents a new problem. Simply applying existing methods to address this (i.e., BDH-MTL) is inadequate. Therefore, we propose a novel method called ADH-MTL, which includes three key designs with the following four design principles. First, in personalized healthcare, building an individual model for each patient can be impractical for new patients. Group-level modeling addresses this challenge by first assigning a new patient to a relevant group and then applying the group-specific model, thereby enhancing applicability for newcomers. Second, the increased complexity inherent in personalized healthcare models poses another major challenge. Matrix decomposition can mitigate this complexity by breaking down high-dimensional matrices (e.g., four-dimensional structures) into multiple lower-dimensional matrices. Third, while Bayesian networks have been utilized in IS research for various tasks, their application in our work is unique: we use them to model dependencies between model parameters and other relevant factors, such as model performance and relation parameters. This joint modeling approach allows the parameter learning process to incorporate potentially relevant factors, resulting in more effective parameter learning. Fourth, explicitly accounting for both the differences and similarities among components in a deep learning model can lead to enhanced performance. In ADH-MTL, we model the relation parameters between these components and construct two separate parameters with the shared priors, ensuring that both differences and similarities are captured. The resultant performance improvement underscores the importance of this principle.

## 6.2 Managerial and Practical Implications

Effective chronic disease management is a lifelong process, encompassing pre-treatment, treatment, and post-treatment. Our study offers benefits at each phase.

In the pre-treatment phase, chronic diseases often develop unnoticed, but timely intervention is critical to prevent further progression. Our expert interviews also highlighted the limitations of traditional self-reported assessments and the benefits of wearable devices. By leveraging wearable sensors and AI, our method proposes a convenient and viable approach for multi-disease assessment. By modeling the relationships among patients and diseases, our method provides more accurate assessment results, enabling patients and their families to become aware of potential disease risks and take appropriate actions, such as visiting a hospital for professional consultation. Hence, our study enables precautionary measures during the pre-treatment phase, which can typically lead to better outcomes compared to waiting until the disease progresses to a severity noticeable by the patient.

In the treatment phase, while a patient usually visits a specific section (e.g., the department of neurology when the patients suspect the existence of Parkinson's disease), doctors should adopt the collaborative approach, considering the comorbid diseases and depression simultaneously. For instance, even if the patients go to the department of neurology, the doctors can use our method to monitor multiple diseases and depression to make an integrated treatment. While professional examination can be taken for a more accurate result, our assessment results can act as a reference. For instance, suggest which examination needs to be taken among various expensive and complicated examinations to further confirm the existence of a certain disease.

In the post-treatment phase, timely monitoring is critical for understanding patient status and evaluating treatment effectiveness. Given the high cost of in-hospital monitoring and the extended duration of chronic disease management, wearable sensor-based remote monitoring has become a practical and cost-effective alternative. This study proposes a novel method that leverages wearable sensors and AI to continuously and efficiently analyze collected data for effective monitoring. Moreover, patients often experience reduced quality of life during long recovery periods, and depression is prevalent among them. Our method not only monitors physical health but also tracks mental health, enabling doctors and patients' families to gain a comprehensive understanding of patient well-being and provide timely interventions or adjust treatment plans if necessary.

**6.3 Limitations and Future Research**

This study has several limitations that point to valuable avenues for future research. First, the proposed method is primarily predictive and focuses on assessing patients' health status. Although this provides important insights for early assessment and continuous monitoring, the model does not prescribe personalized or adaptive intervention strategies to improve patient outcomes. Future research could integrate predictive analytics with prescriptive or recommendation-driven models, such as reinforcement learning–based decision systems or digital therapeutic modules, to form a comprehensive framework for collaborative care. Second, our model primarily leverages wearable sensor data to capture fine-grained behavioral signals. While this provides valuable insights, incorporating multi-device data, such as sleep patterns, daily mobility metrics, and activity levels collected from smartphones or other mobile devices, could significantly enhance the comprehensiveness and accuracy of disease assessment. Finally, this study focuses specifically on depression, which is one of the most prevalent mental health issues among patients with chronic diseases. Future research could extend the current framework to other mental health diseases such as anxiety, stress, sleep disorders, or cognitive decline. Examining whether the proposed method generalizes to these diseases would not only broaden the applicability of the framework but also advance a more comprehensive understanding of mental health assessment in chronic disease management.

## REFERENCES

Abbasi A, Parsons J, Pant G, Sheng ORL, Sarker S (2024) Pathways for design research on artificial intelligence. *Information Systems Research* 35(2):441–459.

Aron R, Pathak P (2021) Disaggregating the differential impact of healthcare IT in complex care delivery: Insights from field research in chronic care. *J Assoc Inf Syst* 22(3):8.

Baird A, Angst C, Oborn E (2020) MIS Quarterly research curation on health information technology research curation team. *MIS Quarterly*.

Bardhan I, Chen H, Karahanna E (2020) Connecting systems, data, and people: A multidisciplinary research roadmap for chronic disease management. *MIS Quarterly* 44(1).

Ben-Assuli O, Padman R (2020) Trajectories of repeated readmissions of chronic disease patients: Risk stratification, profiling, and prediction. *MIS Quarterly* 44(1):201–227.

Blei DM, Kucukelbir A, McAuliffe JD (2017) Variational inference: A review for statisticians. *J Am Stat Assoc* 112(518):859–877.

Brohman K, Addas S, Dixon J, Pinsonneault A (2020) Cascading feedback: A longitudinal study of a feedback ecosystem for telemonitoring patients with chronic disease. *MIS Quarterly* 44(1):421–450.

Chen YQ, Zhang T, Jiang XL, Chen Q, Gao CL, Huang WL (2024) Fedbone: Towards large-scale federated multi-task learning. *J Comput Sci Technol* 39(5):1040–1057.

Chowdhury SR, Das DC, Sunna TC, Beyene J, Hossain A (2023) Global and regional prevalence of multimorbidity in the adult population in community settings: a systematic review and meta-analysis. *EClinicalMedicine* 57.

Crouse JJ, Carpenter JS, Song YJC, Hockey SJ, Naismith SL, Grunstein RR, Scott EM, Merikangas KR, Scott J, Hickie IB (2021) Circadian rhythm sleep-wake disturbances and depression in young people: implications for prevention and early intervention. *Lancet Psychiatry* 8(9):813-823.

Cui, S, Mitra, P (2024) Automated multi-task learning for joint disease prediction on electronic health records. *Advances in Neural Information Processing Systems*, 37, 129187-129208.

Deng L, Yin T, Li Z, Ge Q (2023) Analysis of the Effectiveness of CNN-LSTM Models Incorporating Bert and Attention Mechanisms in Sentiment Analysis of Data Reviews. *2023 4th International Conference on Big Data and Informatization Education (ICBDIE 2023) (Atlantis Press)*, 821–829.

Dirgová Luptáková I, Kubovčík M, Pospíchal J (2022) Wearable sensor-based human activity recognition with transformer model. *Sensors* 22(5):1911.

Fortuin V (2022) Priors in bayesian deep learning: A review. *International Statistical Review* 90(3):563–591.

Ferrari A, Micucci D, Mobilio M, Napoletano P (2023) deep learning and model personalization in sensor-based human activity recognition. *Journal of Reliable Intelligent Environments* 9(1): 27-39.

Fritschi C, Quinn L (2010) Fatigue in patients with diabetes: a review. *J Psychosom Res* 69(1):33-41.

Hacker K (2024) The burden of chronic disease. *Mayo Clin Proc Innov Qual Outcomes* 8(1):112–119.

Hervella ÁS, Rouco J, Novo J, Ortega M (2024) Multi-Adaptive Optimization for multi-task learning with deep neural networks. *Neural Networks* 170:254–265.

Hulshof CM, van der Leeden M, van Netten JJ, Gijssel M, Evers J, Bus SA, Pijnappels M (2024) The association between peripheral neuropathy and daily-life gait quality characteristics in people with diabetes. *Gait Posture* 114:152-159.

Hybels CF, Landerman LR, Blazer DG (2012) Age differences in symptom expression in patients with major depression. *Int J Geriatr Psychiatry* 27(6):601-611.

J. Katon W (2011) Epidemiology and treatment of depression in patients with chronic medical illness. *Dialogues Clin Neurosci* 13(1):7-23.

Jafari M, Shoeibi A, Khodatars M, Ghassemi N, Moridian P, Alizadehsani R, Khosravi A, Ling SH, Delfan N, Zhang YD (2023) Automated diagnosis of cardiovascular diseases from cardiac magnetic resonance imaging using deep learning models: A review. *Comput Biol Med* 160:106998.

Katon WJ, Lin EHB, Von Korff M, Ciechanowski P, Ludman EJ, Young B, Peterson D, Rutter CM, McGregor M, McCulloch D (2010) Collaborative care for patients with depression and chronic illnesses. *New England Journal of Medicine* 363(27):2611-2620.

Kendall A, Gal Y, Cipolla R (2018) Multi-task learning using uncertainty to weigh losses for scene geometry and semantics. *Proceedings of the IEEE conference on computer vision and pattern recognition* 7482–7491.

Li C, Georgiopoulos M, Anagnostopoulos G C (2014) Multitask classification hypothesis space with improved generalization bounds. *IEEE transactions on neural networks and learning systems*, 26(7), 1468-1479.

Lin YK, Chen H, Brown RA, Li SH, Yang HJ (2017) Healthcare predictive analytics for risk profiling in chronic care. *MIS Quarterly* 41(2):473-496.

Lin YK, Fang X (2021) First, do no harm: Predictive analytics to reduce in-hospital adverse events. *Journal of Management Information Systems* 38(4):1122-1149.

Lippi G, Montagnana M, Favaloro EJ, Franchini M (2009) Mental depression and cardiovascular disease: a multifaceted, bidirectional association. *Semin Thromb Hemost* (© Thieme Medical Publishers), 325-336.

Liu X, Zhang B, Susarla A, Padman R (2019) Go to YouTube and call me in the morning: Use of social media for chronic conditions. *Liu, X., Zhang, B., Susarla, A., and Padman*:257-283.

De Mello MT, de Aquino Lemos V, Antunes HKM, Bittencourt L, Santos-Silva R, Tufik S (2013) Relationship between physical activity and depression and anxiety symptoms: a population study. *J Affect Disord* 149(1-3):241-246.

Mezuk B, Eaton WW, Albrecht S, Golden SH (2008) Depression and type 2 diabetes over the lifespan: a meta-analysis. *Diabetes Care* 31(12):2383-2390.

Mohammad-Djafari A (2021) Regularization, Bayesian inference, and machine learning methods for inverse problems. *Entropy* 23(12):1673.

Nunavath V, Johansen S, Johannessen TS, Jiao L, Hansen BH, Berntsen S, Goodwin M (2021) Deep learning for classifying physical activities from accelerometer data. *Sensors* 21(16):5564.

NHS (2023) Symptoms - Depression in adults. Accessed August 26, 2025, https:// www. nhs.uk/mental-health/conditions/depression-in-adults/symptoms/

Nishi K, Kim J, Li W, Pfister H (2024) Joint-task regularization for partially labeled multi-task learning. *Proceedings of the IEEE/CVF Conference on Computer Vision and Pattern Recognition* 16152–16162.

Nutt D, Wilson S, Paterson L (2008) Sleep disorders as core symptoms of depression. *Dialogues Clin Neurosci* 10(3):329-336.

Padmanabhan B, Fang X, Sahoo N, Burton-Jones A (2022) Machine Learning in Information Systems Research. *MIS Quarterly* 46(1).

Pallewar M, Pawar VR, Gaikwad AN (2024) Human Anomalous Activity detection with CNN-LSTM approach. *Journal of Integrated Science and Technology* 12(1):704.

Pirker W, Katzenschlager R (2017) Gait disorders in adults and the elderly: A clinical guide. *Wien Klin Wochenschr* 129(3):81-95.

Rahimian E, Javadi G, Tung F, Oliveira G (2023) DynaShare: Task and instance conditioned parameter sharing for multi-task learning. *Proceedings of the IEEE/CVF Conference on Computer Vision and Pattern Recognition* 4535–4543.

Ruder S (2017) An overview of multi-task learning in deep neural networks. *arXiv preprint arXiv:1706.05098*.

Sen P, Borcea C (2024) Fedmtl: Privacy-preserving federated multi-task learning. *ECAI 2024 (IOS Press)*, 1993–2002.

Sang W, Ding Z, Li M, Liu X, Liu Q, Yuan S (2024) Prestack simultaneous inversion of P-wave impedance and gas saturation using multi-task residual networks. *Acta Geophysica* 72(2):875–892.

Sloman L, Berridge M, Homatidis S, Hunter D, Duck T (1982) Gait patterns of depressed patients and normal subjects. *Am J Psychiatry* 139(1):94-97.

Tanaka T (1998) A theory of mean field approximation. *Adv Neural Inf Process Syst* 11.

Tan YY, Chow CO, Kanesan J, Chuah JH, Lim Y (2023) Sentiment analysis and sarcasm detection using deep multi-task learning. *Wirel Pers Commun* 129(3):2213–2237.

Thomas, D R (2006) A general inductive approach for analyzing qualitative evaluation data. *American journal of evaluation* 27(2), 237-246.

Tsouvalas V, Ozcelebi T, Meratnia N (2025) Many-task federated fine-tuning via unified task vectors. *arXiv preprint arXiv:2502.06376*.

Uddin M A, Talukder M A, Uzzaman M S, Debnath C, Chanda M, Paul S, Aryal, S (2024) Deep learning-based human activity recognition using CNN, ConvLSTM, and LRCN. *International Journal of Cognitive Computing in Engineering*, 5, 259-268.

Viroli C (2012) On matrix-variate regression analysis. *J Multivar Anal* 111:296–309.

Wang X, Zhang R, Zhu X (2024) What can we learn from multimorbidity? A deep dive from its risk patterns to the corresponding patient profiles. *Decis Support Syst* 186:114313.

Wu M, Mirkin S, McPhail MN, Wajeeh H, Nagy S, Florent-Carre M, Blavo C, Beckler MD, Amini K, Kesselman MM (2024) A Comprehensive Review of Lyme Disease: A Focus on Cardiovascular Manifestations. *Cureus* 16(5).

Xie J, Liu X, Dajun Zeng D, Fang X (2022) Understanding Medication Nonadherence from Social Media: A Sentiment-Enriched Deep Learning Approach. *MIS Quarterly* 46(1).

Xie J, Zhang B, Ma J, Zeng D, Lo-Ciganic J (2021a) Readmission prediction for patients with heterogeneous medical history: A trajectory-based deep learning approach. *ACM Transactions on Management Information Systems (TMIS)* 13(2):1-27.

Xie J, Zhang Z, Liu X, Zeng D (2021b) Unveiling the hidden truth of drug addiction: A social media approach using similarity network-based deep learning. *Journal of Management Information Systems* 38(1):166-195.

Xie S, Yu Z, Lv Z (2021) Multi-disease prediction based on deep learning: a survey. *Computer Modeling in Engineering & Sciences* 128(2):489-522.

Yin P, Sun Y, Gao Z, Wang R, Yao Y (2024) MAInt: A multi-task learning model with automatic feature interaction learning for personalized recommendations. *Inf Sci (N Y)* 665:120362.

Yu S, Chai Y, Chen H, Sherman S, Brown R (2022) Wearable sensor-based chronic condition severity assessment: An adversarial attention-based deep multisource multitask learning approach. *MIS Quarterly* 46(3):1355-1394.

Yu S, Chai Y, Samtani S, Liu H, Chen H (2024) Motion sensor-based fall prevention for senior care: A hidden markov model with generative adversarial network approach. *Information Systems Research* 35(1):1-15.

Zhang J, Liu Y, Yuan H (2023) Attention-based residual BiLSTM networks for human activity recognition. *IEEE Access* 11:94173–94187.

Zhang Y, Li X, Rong L, Tiwari P (2021) Multi-task learning for jointly detecting depression and emotion. *2021 IEEE International Conference on Bioinformatics and Biomedicine (BIBM)* (IEEE), 3142-3149.

Zhao Y, Wang X, Luo Y (2023) Research on Human Activity Recognition Algorithm Based on LSTM-1DCNN. *Computers, Materials & Continua* 77(3).

Zhu H, Samtani S, Chen H, Nunamaker Jr JF (2020) Human identification for activities of daily living: A deep transfer learning approach. *Journal of Management Information Systems* 37(2):457-483.